\documentclass[10pt]{article} 
\usepackage[preprint]{tmlr}

\usepackage{enumitem}
\usepackage{lipsum}
\usepackage{hyperref}       
\usepackage{url}            
\usepackage{booktabs}       
\usepackage{amsfonts}       
\usepackage{xcolor}         
\usepackage{amsmath,amsfonts,amsbsy}
\usepackage[mathscr]{euscript}
\usepackage[ruled,linesnumbered,vlined,shortend]{algorithm2e}
\usepackage{algorithmicx}
\usepackage{graphicx}
\usepackage{caption}
\usepackage{subcaption}
\usepackage{booktabs}
\usepackage{multirow}
\usepackage{tabularx}
\usepackage{tablefootnote}
\usepackage{threeparttable}
\usepackage{mdwlist}
\usepackage{makecell}
\usepackage{diagbox}
\usepackage{relsize}
\usepackage{makecell}
\usepackage{hyperref}
\usepackage[T1]{fontenc}
\usepackage{amsthm, amssymb}
\usepackage{comment}
\usepackage{tikz}
\usetikzlibrary{arrows.meta, positioning}
\usetikzlibrary{shapes.geometric}
\usepackage{diagbox}
\usepackage{capt-of}
\usepackage{subcaption}
\usepackage{booktabs}
\usepackage{bbm}
\usepackage{float}

\PassOptionsToPackage{options}{natbib}
\usepackage{natbib}
\usepackage[capitalise,nameinlink]{cleveref}
\crefname{section}{Sec.}{Sec.}
\Crefname{section}{Section}{Sections}
\crefname{listing}{List.}{List.}
\crefname{listing}{Listing}{Listings}
\Crefname{listing}{Listing}{Listings}
\crefname{lstlisting}{Listing}{Listings}
\Crefname{lstlisting}{Listing}{Listings}
\usepackage{colortbl,xcolor,pgf}
\newcommand{\method}[0]{NewPINNs\xspace}

\title{\method: Physics-Informing Neural Networks Using Conventional Solvers for Partial Differential Equations}

\author{\name Satish Chandran\footnotemark[1]
\email schan360@ucr.edu \\
\addr Department of Mathematics, University of California Riverside
\AND
\name Maedeh Makki\thanks{Denotes equal contribution}
\email mmakk004@ucr.edu \\
\addr Department of Mechanical Engineering, University of California Riverside
\AND
\name Maziar Raissi
\email maziarr@ucr.edu \\
\addr Department of Mathematics, University of California Riverside
\AND
\name Adrien Grenier
\email grenier.a.2@pg.com\\
\addr German Innovation Center, Procter \& Gamble Service GmbH
\AND
\name Behzad Mohebbi
\email mohebbi.b@pg.com \\
\addr German Innovation Center, Procter \& Gamble Service GmbH
}

\def\month{MM}  
\def\year{YYYY} 
\def\openreview{\url{https://openreview.net/forum?id=XXXX}} 

\begin{document}

\maketitle

\begin{abstract}
We introduce \method, a physics-informing learning framework that couples neural networks with conventional numerical solvers for solving differential equations. Rather than enforcing governing equations and boundary conditions through residual-based loss terms, \method integrates the solver directly into the training loop and defines learning objectives through solver-consistency. The neural network produces candidate solution states that are advanced by the numerical solver, and training minimizes the discrepancy between the network prediction and the solver-evolved state. This pull–push interaction enables the network to learn physically admissible solutions through repeated exposure to the solver’s action, without requiring problem-specific loss engineering or explicit evaluation of differential equation residuals. By delegating the enforcement of physics, boundary conditions, and numerical stability to established numerical solvers, \method mitigates several well-known failure modes of standard physics-informed neural networks, including optimization pathologies, sensitivity to loss weighting, and poor performance in stiff or nonlinear regimes. We demonstrate the effectiveness of the proposed approach across multiple forward and inverse problems involving finite volume and finite element solvers.
\end{abstract}


\section{Introduction}		\label{sec:intro}
Solving partial differential equations (PDEs) is a fundamental task in many different scientific and engineering disciplines such as electrodynamics, fluid dynamics, and solid mechanics \citep{JacksonEM, panton, dym2013solid, Zheng}. However, outside of simplified toy-models, it is nearly impossible to write down an analytical solution to a given differential equation system \citep{evans10}. As a result, myriad numerical methods such as the finite difference method (FDM), finite volume methods (FVM), finite element methods (FEM), lattice Boltzmann methods (LBM), material point methods (MPM), etc., have been developed over the centuries to solve PDEs accurately for many scientific and engineering problems \citep{BrennerScottFEM,LevequeFDM,LeVequeFVM, LarssonThomeeNumPDE, LarsenFEM, HOSSEINI2023, Ginzburg2008, XLB, DEVAUCORBEIL2020, nguyen2023material, haughey2009boundless}.

Over the past decade deep-learning has been used to solve many scientific and engineering problems involving differential equations. Traditional data-driven neural networks often struggle with physical consistency when modeling complex systems, requiring a prohibitively large number of samples and often failing to generalize beyond training distributions \citep{RaissiHiddenFluid2020}. To address this, the physics-informed neural networks (PINNs) were developed as a flexible method to solve forward and inverse problems \citep{RAISSI2019, ZhaoZhangLouWang2024, GAO2022114502, YANG2025103997}. In the PINNs framework, a neural network learns the solution of a PDE by having its loss function be composed of residuals of the differential equation and the boundary/initial conditions \citep{RAISSI2019}. Despite their success, PINNs face significant structural limitations that hinder their reliability in complex scenarios:

\begin{itemize}
  \item Optimization Difficulties: The standard PINN loss function is a complex weighted sum of PDE residuals, boundary conditions, and initial conditions. This creates a highly non-convex loss landscape that is difficult to optimize, often leading to convergence failures in stiff systems or multi-scale problems \citep{10.5555/3692070.3693785, SONG2024112781}.
  \item Hyperparameter Sensitivity: Balancing the competing terms in the loss function requires tuning multiple scalar weights, a process that is often stiff and problem-dependent.
  \item Spectral Bias: Neural networks tend to learn low-frequency functions first, which causes standard PINNs to struggle with high-frequency or chaotic dynamics where the PDE residual does not provide a strong enough gradient signal for fine-scale features \citep{YE2024111006, app14083204, JIN2024107887}.
\end{itemize}

To address these challenges, many different approaches have been developed. These methods can broadly be classified into 3 distinct categories, namely PINNs, hybrid neural-numerical methods, and data-driven methods.

\paragraph{Physics-Informed Neural Networks (PINNs):}
Since the introduction of ``classical'' PINNs, there have been many modified versions of them to improve their performance across multiple scientific and engineering tasks \citep{RAISSI2019,WANG2024116813, JAGTAP2020113028, LEE2024117000}. Wang et al.  divided their simulation domain into subdomains and trained local PINNs on each one to mitigate very small and very large gradients, and also address poor scaling of PDE residuals \citep{wangetal}.

\paragraph{Hybrid Neural-Numerical Models:}
Combining traditional numerical solvers with neural networks to use the strength of both approaches, has gained significant interest in recent years \citep{Zhang2024, Oommen2024, MERLET2024106316}. The neural ODE model implements the neural network output as a derivative in an ODE solver, and then used an integrator to solve the  ODE and compute the gradients to train the network \citep{NEURIPS2018_69386f6b}. The NNfoil framework combines PINNs with a finite element mesh transformation to simulate subsonic, inviscid flow around airfoils \citep{CaoSongZhang2024}. Time-Stepping-Oriented Neural Network (TSONN) resolves the ill-conditioning in PINNs by adjusting the Jacobian condition number to enhance performance \citep{CaoZhang2025}. Universal Differential Equations (UDEs) split the solution space into aseveral neural networks and feeds them into traditional solvers to train them \citep{Rackauckas}. The neural-initialized Newton (NiN) strategy accelerates nonlinear finite element solvers by using a pretrained physics-informed neural operator as an initial guess for Newton–Raphson iterations \citep{taghikhani2025neuralinitializednewtonacceleratingnonlinear}. The Pretrained Finite Element Method (PFEM) introduces a two-stage hybrid framework in which a transformer-based physics-informed neural operator is trained solely from PDE constraints on unstructured point clouds and subsequently used to warm-start classical finite element solvers \citep{wang2026pretrainfiniteelementmethod}.

\paragraph{Data-Driven Discovery of Differential Equations:}
Extracting physical properties from known or measured data (i.e.\ inverse problems) is another critical problem in many scientific and engineering domains \citep{Egan2024, PU2023107051, MEIDANI2021113831, Xu2023}. The Sparse Identification of Nonlinear Dynamics (SINDy) algorithm is a popular method for discovering governing equations from data by expressing derivatives as sparse combinations of candidate (often polynomial) basis functions of a few dominant terms, and uses sparse regression to identify them from time series data \citep{Huang2022}. However SINDy requires a comprehensive library of functions and often fail to handle noisy data \citep{BRUNTON2016710}. Raissi et al.\ used a PINN-like training approach to find the unknown parameters of equations by approximating the solution and non-linear differential operators as neural networks and fitting the residual to the observed data \citep{MaziarRaissi}. Long et al. developed PDE-Net 2.0, which learns coefficients of PDEs from time dependent data by combining both convolutional and symbolic neural networks \citep{LONG2019108925}.Neural operator methods such as DeepONet and Fourier Neural Operators have also gained traction in recent years. The neural operator frameworks aim to learn global mappings between function spaces, typically by training on large datasets of input–output pairs generated by numerical solvers \citep{Lu2021, fno2020}. While these approaches have demonstrated strong performance in data-rich regimes, they rely on extensive precomputed datasets and primarily function as supervised surrogate models.

To address these fundamental challenges, we introduce \method, a novel framework that fundamentally alters how physics is injected into the network. Instead of forcing the network to minimize a PDE residual derived via automatic differentiation, NewPINNs couples the network with a traditional numerical solver in a ``pull-push'' closed-loop solver-consistency training procedure (see Figure \ref{fig:newpinns1}).

\begin{figure}[ht]
\centering
\begin{tikzpicture}[scale=1, node distance=1.0cm, every node/.style={font=\small}, >=Stealth]

\tikzstyle{bluebox} = [draw, fill=blue!20, minimum width=1.5cm, minimum height=1.0cm]
\tikzstyle{greenbox} = [draw, fill=green!20, minimum width=1.5cm, minimum height=1.0cm]
\tikzstyle{state} = [draw, fill=purple!20, minimum width=1.5cm,minimum height=1.0cm]
\tikzstyle{state2} = [draw, fill=orange!20, minimum width=1.5cm,minimum height=1.0cm]

\node (params) [greenbox] {\shortstack{PDE \\ Parameters}};
\node (nn) [bluebox, right=of params] {\shortstack{Neural\\ Network}};
\node (nn_output) [state, right=of nn] {\,\,\,$\widehat{u}(x,t_n)$\,\,\,\,};
\node (solver) [bluebox, right=of nn_output] {\shortstack{Numerical\\ Solver}};
\node (solver_output) [state2, right=of solver] {$u_{S}(x,t_{n+1})$};
\node (loss) [bluebox, right=of solver_output] {Loss};
\node (nn_output2)  [state, above=of nn_output] {$\widehat{u}(x,t_{n+1})$};
\node (gd)  [bluebox, below=of solver] {\shortstack{Optimizer}};

\draw[->] (params) -- (nn);
\draw[->] (nn) -- (nn_output);
\draw[->] (nn_output) -- (solver);
\draw[->] (solver) -- (solver_output);
\draw[->] (solver_output) -- (loss);
\draw[->] (nn.north) -| ++(0,1.5) |- (nn_output2.west);
\draw[->] (nn_output2.east) -| ++(7.31,-1) -| (loss.north);
\draw[->] (loss.south) |- ++(0,-1.5) |- (gd.east);
\draw[->] (gd.west) |- ++(-4.0,0) -| (nn.south);

\end{tikzpicture}
\caption{Transient \method: At each training iteration, the neural network produces a candidate solution at time $t_n$ which is advanced to $t_{n+1}$ by the numerical solver. A solver-consistency loss compares this solver-evolved state with the network’s prediction at the next time step, enforcing temporal consistency under solver evolution. The solver is treated as a black-box operator and does not participate in gradient computation.}
\label{fig:newpinns1}
\end{figure}

In this framework, the neural network acts as a generator of candidate solution states (``pull''), while the external solver defines a numerical evolution operator that advances these states according to a prescribed discretization (``push''). The loss function is then defined as a solver-consistency objective, which enforces agreement between the network’s prediction and the result of applying the solver to that prediction. This approach offers several distinct advantages over standard PINNs:

\begin{itemize}
  \item Simplified Optimization: By removing PDE residuals and boundary constraints from the loss function, we eliminate the conflict between competing loss terms and the need for delicate hyperparameter tuning.

  \item Solver Agnosticism: Unlike ``differentiable physics'' approaches that require specialized differentiable solvers, NewPINNs treats the solver as a black box. This allows the use of established, highly optimized commercial solvers (e.g., COMSOL, Ansys) and does not require the solver to be differentiable.

  \item Surrogate Modeling: By enforcing consistency under solver evolution, the network is trained to produce physically admissible states that respect the stability and invariance properties of the underlying numerical discretization. Thus the neural network effectively learns to become a fast surrogate (or ``twin'') of the solver, capable of predicting complex, non-linear, and even chaotic dynamics that standard PINNs can often fail to capture.
\end{itemize}

Viewed from an operator perspective, \method does not attempt to learn a global mapping from inputs to solutions using fixed solver-generated datasets. Instead, it enforces physical consistency by minimizing a solver-induced operator residual evaluated on the network’s own predictions. The solver does not act as a teacher or data generator, but rather as a constraint operator whose action defines admissible solutions through invariance or consistent temporal evolution. This distinction allows \method to operate effectively in data-scarce regimes and to leverage legacy numerical solvers as black-box components, while maintaining a principled notion of physical validity without the need for explicit PDE residual losses.

The source code is available on \url{https://github.com/chandran-satish/NewPINNs}.

\section{Methodology}		\label{sec:method}
In general, we are interested in solving partial differential equations (PDEs) of the form
\begin{equation}
\begin{aligned}
    &\frac{\partial u}{\partial t}= \mathcal{N}[u(x,t);\alpha] \,\, \forall (x,t) \in \Omega \times [0,T] \\
    &\mathcal{I}[u(x,0)] = 0\,\,\,\,\, \forall x \in \Omega\\
    &\mathcal{B}[u(x,t)] = 0\,\,\,\,\forall (x,t) \in \partial\Omega \times [0,T]
\end{aligned}
\label{generalpde}
\end{equation}
where $\mathcal{N}$ is a non-linear differential operator parameterized by $\alpha$,  $\Omega$ is the spatial domain in $\mathbb{R}^d$ and $\partial \Omega$ the corresponding boundary, $\mathcal{B}$ and $\mathcal{I}$ denote the sets of boundary and initial conditions respectively, and $u$ is the unknown solution function.
\subsection{Physics-Informed Neural Networks (PINNs)}
Physics-Informed Neural Networks (PINNs) aim to approximate $u$ using neural networks by incorporating the PDE directly into the loss function. During training, the network is evaluated at a set of collocation points in the spatio-temporal domain, and automatic differentiation is used to compute the residuals of the PDE and auxiliary constraints. These residuals form the basis of the loss function that drives the optimization of \( \theta \).

The standard PINN loss function is composed of multiple terms, corresponding to the PDE constraint, boundary condition (BC) constraint, or initial condition (IC) constraint
\begin{equation}
    \begin{aligned}
    \mathcal{L}(\theta) =  \lambda_{PDE}\mathcal{L}_{\text{PDE}} + \lambda_{BC}\mathcal{L}_{\text{BC}} + \lambda_{IC}\mathcal{L}_{\text{IC}},
    \end{aligned}
\end{equation}
where each term measures the mean squared residual at (randomly sampled) collocation points which are given by
\begin{equation}
\begin{aligned}
    \mathcal{L}_{\text{PDE}} &= \frac{1}{N_{\text{PDE}}} \sum_{i=1}^{N_{\text{PDE}}} \bigg|\bigg| \frac{\partial}{\partial t}\widehat{u}(x_i, t_i; \theta) - \mathcal{N}[\widehat{u}(x_i, t_i; \theta)] \bigg|\bigg|^2 \text{ with } x_i \in \Omega,
\end{aligned}
\end{equation}
\begin{equation}
\begin{aligned}
    \mathcal{L}_{\text{BC}} = \frac{1}{N_{\text{BC}}} \sum_{j=1}^{N_{\text{BC}}} \bigg|\bigg| \mathcal{B}[\widehat{u}(x_j, t_j; \theta)]  \bigg|\bigg|^2 \text{ with } x_j \in \partial\Omega,
\end{aligned}
\end{equation}
and
\begin{equation}
\begin{aligned}
    &\mathcal{L}_{\text{IC}} = \frac{1}{N_{\text{IC}}} \sum_{k=1}^{N_{\text{IC}}} \bigg|\bigg| \mathcal{I}[\widehat{u}(x_k, t_0; \theta)]  \bigg|\bigg|^2 \text{ with } x_k \in \Omega,
\end{aligned}
\end{equation}
where $N_{PDE}$, $N_{BC}$, and $N_{IC}$ refer to the total number of collocation points sampled from the domain for each residual term respectively. The weights $\lambda_{PDE}$,  $\lambda_{BC}$,  and $\lambda_{IC}$ allow for the tuning the relative importance of each term in the loss function. This formulation enables PINNs to unify data-driven learning and physics-based modeling. However it also introduces challenges, such as sensitivity to the choice of weights, difficulty in training for stiff PDEs, and potential imbalance between residual components, especially in multi-scale or convection-dominated flows \citep{Ji2021}.

\subsection{\method: Physics-Informing Neural Networks using Numerical Solvers }
The \method approach differs from the PINNs framework in that it directly uses traditional numerical solvers to guide the training of the neural network. Generally, numerical solvers work by discretizing the spatial and temporal domains and compute the approximate solution at every discrete time step \citep{Blazek2015, LeVeque2007}. Solution over the entire domain is obtained by forward step-by-step iterations starting from the initial time $t_0$ and initial conditions. In mathematical terms, we say a solver $\mathcal{S}$ takes in an initial state $u(x,t_0)$, along with boundary conditions $\mathcal{B}[u]$ and a set of physical parameters $\alpha$, and pushes it forward to time $t_{N_{s}}$ in discrete time steps i.e.,
\begin{equation}
\begin{aligned}
   u(x,t_{N_{s}}) = \mathcal{S}[u(x,t_0), \mathcal{B}[u], \alpha, N_s].
\end{aligned}
\label{equ_solver}
\end{equation}
Here, $N_s$ is the number of required iterations inside the solver to compute $u(x,t_{N_{s}})$ from $u(x,t_0)$.  For many PDEs, we can also view this solver mapping as an approximation of the semi-group operator of the system \citep{Chen2023, Tsai2018}. One important note here is that the spatial domain can be discretized in a variety of ways depending on the numerical solver of choice. For example finite-difference solvers will discretize the spatial domain into discrete grid points, while finite element solvers will discretize the spatial domain as a mesh which can take on many complex geometries \citep{Bathe2006}. Thus going forward, we omit the discrete indices of the spatial domain for readability.
\subsubsection{Forward Problems}
In forward problems, we are interested in finding the solution function $u$, given the PDE, the initial and boundary conditions, and the parameters of the equation (i.e., $\alpha$).
\paragraph{\method for Transient Systems:}
Most traditional numerical solvers for time-dependent PDEs (e.g., finite difference methods, finite element methods, or finite volume methods) do not explicitly incorporate continuous time $t$ in their formulations. Instead, they work with discrete time steps $\Delta t$ and a total number of time steps $N_t$, and iteratively advance the solution  from one step to the next \citep{gustafsson2013time}. In this regard, and according to the Eq.\ \eqref{equ_solver}, traditional solvers can accept a value as initial condition—meaning the neural network $f_{NN}$ can provide its output to the solver (pull mechanism). The solver then applies the boundary conditions $\mathcal{B}[u]$, and physical properties $\alpha$ to compute the next step solution by taking the given $N_s$ iterations, and returns its computed value to the neural network (push mechanism) i.e.\
\begin{equation}
\begin{aligned}
   u(x,t_{N_s}) = \mathcal{S}[f_{NN}(t_0), \mathcal{B}[u], \alpha, N_s].
\end{aligned}
\label{equ_solver_pul_push}
\end{equation}
The main idea for transient \method is to treat each temporal grid point as a virtual initial time, and feed the physical parameters of our system and the temporal grid point into our neural network which outputs the approximate PDE solution along the spatial domain i.e.\
\begin{equation}
\begin{aligned}
   \widehat{u}(x, t_{n};\theta)= f_{NN}(t_n,\alpha;\theta).
\end{aligned}
\end{equation}
Then, the numerical solver is applied to each $\widehat{u}(x, t_{n};\theta)$ to push it forward in time by one temporal step to obtain $u(x, t_{n+1})$ i.e.\ 
\begin{equation}
\begin{aligned}
   u_{\text{solver}}(x,t_{n+1}) =  \mathcal{S}[\widehat{u}(x, t_n;\theta), \mathcal{B}[u], \alpha, N_s].
\end{aligned}
\end{equation}
Here, we use the index $n$ to denote the temporal sequence. In a real experiment, it takes $N_s$ steps for the solver to compute the output from the network prediction. The solver output is then compared with the neural network output. This results in the loss function
\begin{equation}
\begin{aligned}
  \mathcal{L}_{\text{solver}}(\theta) = \frac{1}{N_\alpha}\sum_{\alpha}\sum_{n=0}^{N_t-1} \bigg|\bigg| f_{NN}( t_{n+1},\alpha;\theta) - \mathcal{S}[\widehat{u}(x, t_n;\theta), \mathcal{B}[u], \alpha, N_s]  \bigg| \bigg|_{L^2(\Omega)}^2.
\end{aligned}
\end{equation}
Here $\left|\left|\cdot \right| \right|_{L^2(\Omega)}$ refers to the (discrete) $L^2$ norm over the spatial domain $\Omega$. Another important aspect of transient systems is that they are not invariant to initial states unlike steady-state systems, i.e.\ different initial conditions $\mathcal{I}[u(x,t_0)]$ in Eq.\ \eqref{generalpde} will result in different spatiotemporal trajectories for the system. As a result, it is also critical to enforce the initial conditions of the system for the neural network predictions. This is achieved by adding the corresponding loss term
\begin{equation}
\begin{aligned}
   \mathcal{L}_{\text{IC}}(\theta) = \frac{1}{N_\alpha}\sum_{\alpha}\bigg|\bigg| \mathcal{I}[f_{NN}(t_0, \alpha)] \bigg| \bigg|_{L^2(\Omega)}^2.
\end{aligned}
\end{equation}
This results in the total loss function being 
\begin{equation}
\begin{aligned}
   \mathcal{L}_{\text{transient}}(\theta) = \mathcal{L}_{\text{IC}}(\theta) + \mathcal{L}_{\text{solver}}(\theta).
\end{aligned}
\end{equation}
The combination of these loss functions ensure strict adherence to the specified initial conditions, and that the network’s solution remains consistent with the solver’s output for the spatio-temporal trajectories. Minimizing the proposed $\mathcal{L}_{\text{transient}}(\theta)$ results in a neural network which can output the PDE solution along the specified spatiotemporal grid points. The pseudo-code for training the neural network for transient systems is given in Algorithm (\ref{alg2}). We consider neural architectures that either (i) output a discretized field over $\Omega$ directly (e.g.\ U-Nets), written as $f_{\mathrm{NN}}(t,\alpha;\theta)$, or
(ii) output pointwise solution values for given spatiotemporal coordinates
$(x,t)$, written as $f_{\mathrm{NN}}(x,t,\alpha;\theta)$, which are then evaluated over
a spatial grid to form a state vector. For notational simplicity, we refer to both cases using the unified notation
$f_{\mathrm{NN}}(t,\alpha;\theta)$.

\begin{algorithm}[H]
    \footnotesize 
    \DontPrintSemicolon
    \caption{Training Routine for Transient \method}
    \label{alg2}
    \SetKwInOut{Input}{Input}
    \SetKwInOut{Output}{Output}
    \SetKwBlock{Function}{function \normalfont{TransientNewPINNs}$(N_s, \alpha_{\text{train}}, \mathcal{B}[u], M)$}{end}

    \Input{Training set of physical parameters $\alpha_{\text{train}}$, boundary conditions $\mathcal{B}[u]$, initial conditions $\mathcal{I}[u(x,t_0)]$, number of training epochs $M$, learning rate $\eta$}
    \Output{ Trained network parameters $\theta$}

    \Function{
        \For{$n \gets 0$ \KwTo $M-1$}{                              
            $\mathcal{L}(\theta) \gets 0$ \BlankLine
            \For{$\alpha \in \alpha_{\text{train}}$}{
                $\mathcal{L}_{IC} \gets \left\|  \mathcal{I}[f_{NN}(t_0,\alpha)] \right\|^2$ \tcp*{Initial Condition Loss}\BlankLine
                $\mathcal{L}(\theta) \gets \mathcal{L}(\theta) + \mathcal{L}_{IC}$
                
                \For{$k \gets 0$ \KwTo $N_t-1$}{                      
                    $\widehat{u}(x,t_{k}, \alpha; \theta) \gets f_{\text{NN}}(t_{k}, \alpha; \theta)$
                    \tcp*{Network predicts current state}
                    
                    $u_{\text{solver}}(x,t_{k+1}, \alpha) \gets \mathcal{S}[\widehat{u}, \alpha, \mathcal{B}[u]]$
                    \tcp*{Solver pushes to next state}
                    
                    $\mathcal{L}(\theta) \gets \mathcal{L}(\theta) + \left\| f_{\text{NN}}( t_{k+1}, \alpha; \theta) - u_{\text{solver}}(x,t_{k+1}, \alpha) \right\|^2$ 
                }
            }
            $\theta \gets \theta - \eta \nabla_{\theta} \mathcal{L}$
        }
    }
\end{algorithm}
\paragraph{\method for Steady-State Systems:}
For many physical systems, it is known that the PDE will converge to an equilibrium (i.e.\ steady-state) solution $u_{eq}(x)$ as $t$ approaches infinity \citep{Boussaid, LeVeque2007}. In the numerical setting, this means there is some sufficiently large number of solver iterations $N_{\infty}$ that will ensure this, i.e.\
\begin{equation}
\begin{aligned}
   u_{eq}(x) = \mathcal{S}[u(x,t_0), \mathcal{B}[u], \alpha, N_{\infty}].
\end{aligned}
\end{equation}
Of course, if the equilibrium solution is fed into the solver then it will return the equilibrium solution again i.e.\ 
\begin{equation}
\begin{aligned}
   u_{eq}(x) = \mathcal{S}[u_{eq}(x), \mathcal{B}[u], \alpha, N], \,\, \forall \,N > 0,
\end{aligned}
\end{equation}
for any number of iterations $N$ \citep{marwah2023deep,Granas2003}. The key idea behind the \method framework for steady state systems is that we can exploit this fixed/equilibrium point to help train the neural network. We want to train a neural network $f_{NN}(\alpha;\theta)$ that takes in the physical parameters $\alpha$ as an input and outputs $u_{eq}(x)$ directly. To derive the \method loss function, we can take the output of the neural network, and then feed it as an input to the solver and run the solver for a small number of iterations $N_s < < N_{\infty}$. If the neural network is accurate, the mean-squared error should be close to zero. In mathematical terms, we define the \method loss function as 
\begin{equation}
\begin{aligned}
  \mathcal{L}(\theta) = \frac{1}{N_{\alpha}}\sum_{\alpha} \bigg|\bigg|f_{NN}(\alpha;\theta) - \mathcal{S}[f_{NN}(\alpha;\theta), \mathcal{B}[u], \alpha, N_s] \bigg|\bigg|_{L^2(\Omega)}^2
\end{aligned}
\end{equation}
where $\alpha$ is the set of physical parameters we are interested in training over and $N_{\alpha}$ is the number of physical parameter samples. The pseudo-code for training the neural network in steady state systems is given in Algorithm (\ref{alg1}).

\begin{algorithm}[H]
    \footnotesize 
    \DontPrintSemicolon
    \SetKwInOut{Input}{Input}
    \SetKwInOut{Output}{Output}
    \SetKwBlock{Function}{function \normalfont{SteadyStateNewPINNs}$(N_s, \alpha_{\text{train}}, \mathcal{B}[u], M)$}{end}

    \Input{ Number of solver iterations $N_s$, training set of physical parameters $\alpha_{\text{train}}$, boundary conditions $\mathcal{B}[u]$, number of training epochs $M$, learning rate $\eta$}
    \Output{ Trained network parameters $\theta$}

    \Function{
        \For{$n \gets 0$ \KwTo $M-1$}{                                      
            $\mathcal{L}(\theta) \gets 0$                                    \tcp*{Initialize epoch loss}

            \For{$\alpha \in \alpha_{\text{train}}$}{                        
                $\widehat{u}(x, \alpha; \theta) \gets f_{\text{NN}}(\alpha; \theta)$ 
                \tcp*{Get neural network prediction}
                
                $u_{\text{solver}}(x, \alpha) \gets \mathcal{S}[\widehat{u}, \alpha, \mathcal{B}[u], N_s]$ 
                \tcp*{Feed into numerical solver}
                
                $\mathcal{L}(\theta) \gets \mathcal{L}(\theta) + \left\| \widehat{u}(x, \alpha; \theta) - u_{\text{solver}}(x, \alpha) \right\|^2$ 
                \tcp*{Compute mean squared error}
            }
            $\theta \gets \theta - \eta \nabla_{\theta} \mathcal{L}$          \tcp*{Perform gradient descent update}
        }
    }
    \caption{Training Routine for Steady-State \method}
    \label{alg1}
\end{algorithm}
The critical idea here is that by coupling the neural network and numerical solver in training, the neural network is trained to become a ``surrogate/twin'' of the numerical solver. This coupling allows the solver to ``pull'' the neural network into the correct physical solution space. It is also important to note here that unlike PINNs which have explicit loss terms for the boundary conditions, here the boundary conditions are directly enforced by the numerical solver itself.

\paragraph{\method for Optimization-Based Steady-State Systems:}
The steady-state formulation above implicitly assumes that the solver $\mathcal{S}$ drives an initial state toward equilibrium through repeated time-stepping or fixed-point iterations, so that the equilibrium is recovered as a long-time limit. For a large class of steady physical problems (especially in solid mechanics), the equilibrium solution is more naturally characterized as the minimizer of an energy functional, or equivalently as the solution of a (possibly nonlinear) algebraic system obtained from a variational principle, rather than as the limit of a transient process \citep{Bathe2006}. The key observation is that the \method machinery carries over to this setting essentially unchanged, provided the solver $\mathcal{S}$ is reinterpreted as an \emph{iterative optimizer} rather than a time integrator. Consider a steady problem whose solution minimizes a parameterized energy functional
\begin{equation}
\begin{aligned}
   u_{eq}(x) = \operatorname*{arg\,min}_{u \in V_{\mathcal{B}}} \; \Pi_{\alpha}(u),
\end{aligned}
\end{equation}
where $V_{\mathcal{B}}$ is the space of admissible fields satisfying the (Dirichlet) boundary conditions $\mathcal{B}[u]$, and $\Pi_{\alpha}$ is an energy functional parameterized by the physical parameters $\alpha$. As a concrete example, for linear elastostatics on a domain $\Omega$, the displacement field minimizes the total potential energy
\begin{equation}
\begin{aligned}
   \Pi_{\alpha}(u) = \tfrac{1}{2}\, a_{\alpha}(u,u) - \ell_{\alpha}(u),
   \qquad
   a_{\alpha}(u,v) = \int_{\Omega} \sigma_{\alpha}(u) : \varepsilon(v)\, dx,
\end{aligned}
\end{equation}
where $\varepsilon(\cdot)$ is the symmetric strain operator, $\sigma_{\alpha}$ the stress tensor determined by the material parameters $\alpha$ (e.g.\ the Lam\'e coefficients), and $\ell_{\alpha}$ the work of the body forces and surface tractions. The associated first-order optimality (Euler--Lagrange) condition is the discrete equilibrium system
\begin{equation}
\begin{aligned}
   G_{\alpha}(u_{eq}) := \nabla_{u}\Pi_{\alpha}(u_{eq}) = A(\alpha)\,u_{eq} - b(\alpha) = 0,
\end{aligned}
\end{equation}
where $A(\alpha)$ is the (symmetric positive-definite) stiffness operator and $b(\alpha)$ the load vector assembled by the finite element discretization.

In this setting, the role of the solver is played by an iterative optimizer: starting from an initial guess $u^{(0)}$, the optimizer applies $N_s$ update steps that monotonically reduce the energy (or its gradient) and returns the resulting iterate,
\begin{equation}
\begin{aligned}
   u^{(N_s)} = \mathcal{S}[u^{(0)}, \mathcal{B}[u], \alpha, N_s].
\end{aligned}
\end{equation}
Representative choices include preconditioned Krylov methods (e.g.\ conjugate gradients applied to the linear system $A(\alpha)u = b(\alpha)$), Newton or quasi-Newton iterations for nonlinear problems, and projected-gradient schemes for constrained problems such as contact. The minimizer $u_{eq}$ is the unique fixed point of this optimizer. Once the optimality residual $G_{\alpha}(u_{eq})$ vanishes, further iterations leave the state unchanged, exactly as in the time-marching case,
\begin{equation}
\begin{aligned}
   u_{eq}(x) = \mathcal{S}[u_{eq}(x), \mathcal{B}[u], \alpha, N], \quad \forall\, N > 0.
\end{aligned}
\end{equation}
Because this fixed-point structure is identical to that of the iterative steady-state solver, the \method loss is unchanged,
\begin{equation}
\begin{aligned}
   \mathcal{L}(\theta) = \frac{1}{N_{\alpha}}\sum_{\alpha} \bigg|\bigg| f_{NN}(\alpha;\theta) - \mathcal{S}\big[f_{NN}(\alpha;\theta), \mathcal{B}[u], \alpha, N_s\big] \bigg|\bigg|_{L^2(\Omega)}^2,
\end{aligned}
\end{equation}
with the sole reinterpretation that $\mathcal{S}$ now denotes a small number $N_s$ of optimizer iterations applied to the network's own prediction. The network $f_{NN}(\alpha;\theta)$ is thereby trained to produce fields that are stationary under the optimizer, i.e.\ that approximately satisfy the discrete equilibrium equations across the family of physical parameters $\alpha$.

This optimization-based view also clarifies the meaning of the \method residual. For a single (preconditioned) gradient step with preconditioner $P$ and step size $\tau$, the update operator reads $T_{\alpha}(u) = u - \tau\, P^{-1} G_{\alpha}(u)$, so the steady-state residual satisfies
\begin{equation}
\begin{aligned}
   R(u) = u - \mathcal{S}[u, \mathcal{B}[u], \alpha, N_s] \;\propto\; P^{-1} G_{\alpha}(u) = P^{-1}\nabla_{u}\Pi_{\alpha}(u).
\end{aligned}
\end{equation}
Minimizing the \method loss therefore drives the (preconditioned) energy gradient toward zero, directly enforcing first-order optimality of the predicted field. Two practical consequences follow. First, the convergence and stability of training inherit those of the chosen optimizer. A well-conditioned update (e.g.\ conjugate gradients with a robust preconditioner) yields a contractive operator $T_{\alpha}$ with a small contraction factor, whereas a poorly scaled fixed-step iteration may fail to be contractive on stiff or ill-conditioned systems and destabilize training. Second, because discrepancies are measured through the energy gradient, the scale of the network output must be matched to the scale of the equilibrium solution. A large mismatch produces update residuals that are either negligible or overwhelming, leading to degenerate training. As in the time-marching case, the boundary conditions $\mathcal{B}[u]$ are enforced directly by the solver via the admissible function space $V_{\mathcal{B}}$ and the assembled system.

\subsubsection{Inverse Problems}
In this section we focus on the methodology for inverse problems under the \method framework. The goal is to discover unknown  physical parameter variables that satisfy Eq.\ \eqref{generalpde} from  noisy observed data $u_{\text{ob}}(x,t)$. To achieve this, we introduce a novel training methodology that couples the forward and inverse components. The top part of this loop structure has the forward component at top as derived from the approach we discussed in section 2.2.1 to solve Eq.\ \eqref{generalpde} which generalizes the neural network solution over a range of physical parameters $\alpha$. The bottom of the loop structure then uses the same neural network from the forward part to solve the inverse problem. The inverse component of the loop targets the recovery of defined learnable variables $\beta$ by minimizing the inverse loss, given as
\begin{equation}
\begin{aligned}
  \mathcal{L}(\beta) = \frac{1}{N_t}\sum_{i=0}^{N_t - 1}\bigg|\bigg|f_{NN}(t,\beta;\theta) - u_{\text{ob}}(x,t) \bigg|\bigg|_{L^2(\Omega)}^2.
\end{aligned}
\end{equation}
The inverse component of the loop benefits from the forward component's learning of the underlying physics. The external solver remains non-differentiable; the inverse step differentiates only through the neural surrogate $f_{NN}(\cdot;\theta)$ whose parameters have been constrained by solver-consistency training. The pseudo-code for solving inverse problems is given in Algorithm (\ref{alg3}).

\begin{algorithm}
    \footnotesize 
    \DontPrintSemicolon
    \SetKwInOut{Input}{Input}
    \SetKwInOut{Output}{Output}
    \SetKwBlock{Function}{function \normalfont{InverseNewPINNs}$(N_s, \alpha_{\text{train}}, \mathcal{B}[u], M)$}{end}

    \Input{Training set of physical parameters $\alpha_{\text{train}}$, boundary conditions $\mathcal{B}[u]$, initial conditions $\mathcal{I}[u(x,t_0)]$, initial guess for learnable physical parameter $\beta$, number of training epochs $M$, learning rates $\eta_{\text{fw}}$ and $\eta_{\text{inv}}$, observed data $u_{\text{ob}}$}
    \Output{Trained network parameters $\theta$, learnable parameter $\beta$}

    \Function{
        \For{$n \gets 0$ \KwTo $M-1$}{                                      
        
            $\mathcal{L}(\theta) \gets 0$ 
            
            \vspace{-0.2cm}
            \hspace*{0.6cm}{\large\textbf{$\vdots$}} \tcp*{Execute Transient \method training loop}
            \vspace{0.1cm}

            $\theta \gets \theta - \eta_{\text{fw}} \nabla_{\theta} \mathcal{L}$ \tcp*{Update $\theta$ using forward physics loss} 
            \vspace{0.3cm}
            
            \tcp{Inverse Step}
            $\mathcal{L}(\beta) \gets 0$
            
            \For{$k \gets 1$ \KwTo $N_t-1$}{                                
                $\widehat{u}(x,t_{k}, \beta; \theta) \gets f_{\text{NN}}(t_{k}, \beta; \theta)$
                \tcp*{Get neural network prediction}
                
                $\mathcal{L}(\beta) \gets \mathcal{L}(\beta) + \left\| \widehat{u}(x,t_{k}, \beta; \theta) - u_{ob}(x,t_{k})\right\|^2$ \tcp*{compute inverse loss}
            }

            $\beta \gets \beta - \eta_{\text{inv}}\nabla_{\beta}\mathcal{L}(\beta)$
            \tcp*{Update the learnable parameter $\beta$}
        }
    }
    \caption{Training Routine for Inverse \method}
    \label{alg3}
\end{algorithm}

\subsection{Relation to Distillation and Solver-Coupled Learning} \label{sec: distillation}

Although \method leverages numerical solvers during training, it is fundamentally different from supervised surrogate modeling approaches in which a neural network is trained to regress directly onto solution fields generated by a solver. In supervised surrogates, the solver provides fixed labeled data, and the network minimizes a discrepancy between its output and precomputed solver trajectories. NewPINNs does not rely on such offline datasets or static targets. Instead, the numerical solver defines a physics-based consistency operator that evaluates the validity of the network’s own predictions. During training, the neural network produces a candidate solution, which is reinserted into the solver and advanced forward in time or iteration. The loss penalizes inconsistencies between the network output and the solver response to that same output. As a result, the solver does not act as a static teacher but as a dynamic, state-dependent operator whose output depends explicitly on the network prediction. This closed-loop formulation ensures that the training signal is governed by physical consistency rather than by direct imitation of solver-generated data.

\method is also distinct from teacher–student distillation and teacher forcing strategies commonly used in machine learning. In distillation, a student network is trained to reproduce fixed outputs produced by a teacher model, and the teacher behavior is independent of the student’s predictions. In contrast, the solver output in \method depends directly on the network’s current state, and no fixed teacher labels are provided. Furthermore, unlike teacher forcing methods for sequential models, where ground-truth states are injected during training to stabilize optimization, \method preserves the closed-loop feedback between successive predictions and solver evaluations. The network is always trained on its own outputs passed through the solver. This ensures that learning enforces fixed-point consistency in steady-state systems and local semi-group consistency in transient systems. Consequently, \method differs fundamentally from both supervised distillation and sequential training heuristics, and instead learns solutions that are constrained by the underlying physics encoded in the numerical solver.

The steady-state formulation of \method is also related to deep equilibrium models and fixed-point-based learning approaches, which characterize solutions as equilibria of implicit operators \citep{bai2019deepequilibriummodels, bai2020multiscaledeepequilibriummodels, winston2021monotoneoperatorequilibriumnetworks, marwah2023deep}. In NewPINNs, the numerical solver itself defines the fixed-point operator, and training minimizes the discrepancy between the network output and the solver-applied network output. Unlike deep equilibrium networks, which introduce learned implicit layers and require specialized techniques for stability and gradient computation, NewPINNs relies on the numerical solver iterations to define the equilibrium structure. This shifts the burden of stability and convergence from the neural network architecture to the numerical solver, which often have strong theoretical guarantees. Consequently, NewPINNs provides a physics-driven fixed-point formulation that preserves the interpretability and robustness of classical solvers while benefiting from the expressive power of neural networks.


\section{Experiments}		\label{sec:experiments}
In this section we demonstrate the \method framework for across different solver types, namely the finite volume and the finite element method. All experiments use one or two NVIDIA A6000 GPUs unless otherwise specified.

\subsection{The Two-dimensional Fokker--Planck Equation using the Finite Volume Method}
We demonstrate the applicability of \method to the two-dimensional Fokker--Planck equation, a prototypical convection--diffusion equation that arises in statistical physics and stochastic processes as the evolution equation for probability densities under diffusion and drift in an external potential.

On the unit square domain $\Omega = [0,1]^2$ the density $u(x,y,t)$ evolves according to the Fokker-Planck equation given by
\begin{equation}
\frac{\partial u}{\partial t}
=
\nabla \cdot \left( \nabla u + u \nabla V \right),
\qquad (x,y) \in \Omega,
\label{eq:fpe_transient}
\end{equation}
with a parameterized confining potential $V(x,y) = \alpha \sin(2\pi x)\sin(2\pi y)$ for $\alpha \in [1,2]$. No-flux boundary conditions are imposed on $\partial\Omega$, which conserves the total probability. For confining potentials of this form the system relaxes to a steady-state equilibrium given by the Boltzmann density
\begin{equation}
u_{\mathrm{eq}}(x,y)
=
\frac{1}{Z}\exp\!\left(-V(x,y)\right),
\qquad
Z = \int_{\Omega} \exp(-V(x,y))\,dx\,dy,
\label{eq:boltzmann}
\end{equation}
which we use only for evaluation and error analysis.

Reference solutions are generated with a finite volume method (FVM) implemented in FiPy, which enforces the governing equation in integral form over a collection of control volumes \cite{FiPy:2009}. The domain is discretized on a uniform Cartesian grid of $32 \times 32$ square cells, and cell averages of $u$ are evolved in time by balancing diffusive and convective fluxes across the cell faces. Diffusive fluxes use second-order central differences, while the drift term $u\nabla V$ uses an exponential fitting scheme that improves stability in convection-dominated regimes and preserves positivity of the density. Time integration follows a semi-implicit scheme with step size $\Delta t = 1/32$, and each solver call during training advances the system for $N_s = 10$ steps, which substantially reduces transients while remaining inexpensive. The conservative structure of the discretization keeps mass conserved up to solver tolerances throughout training.

\begin{figure}[htbp]
    \centering
    \captionsetup{font=small,labelfont=small}
    \includegraphics[width=0.48\textwidth]{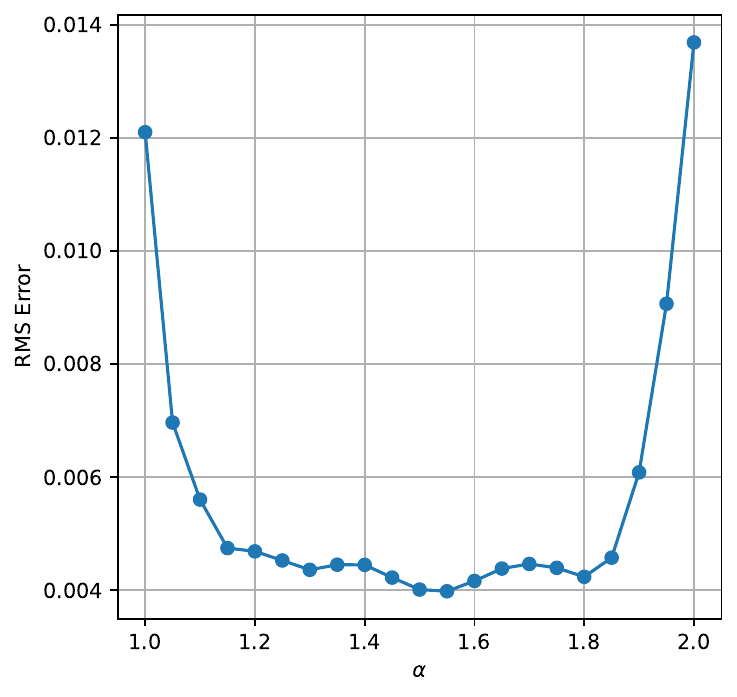}
    \hfill
    \includegraphics[width=0.48\textwidth]{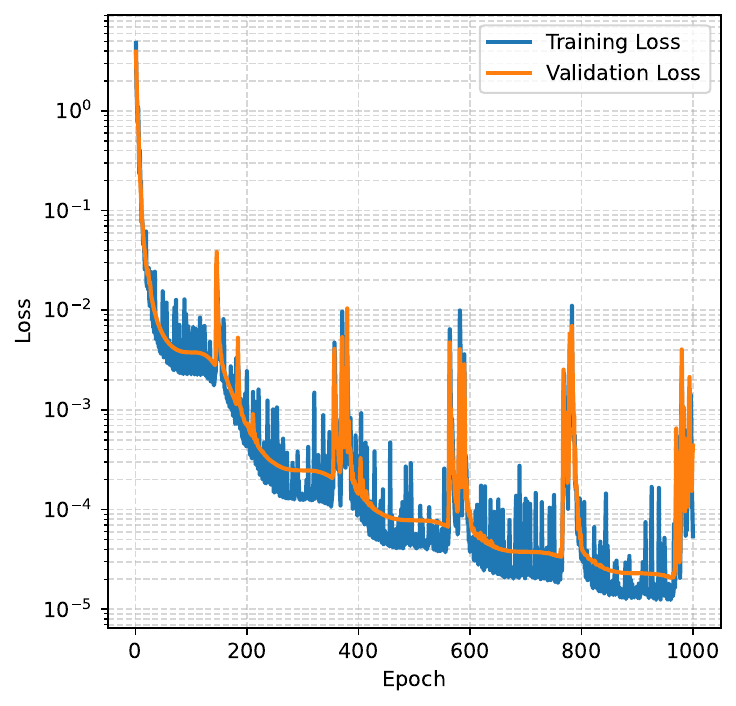}
    \caption{$L^2$ error of the predicted Fokker--Planck solution against the analytical Boltzmann equilibrium as a function of $\alpha$ (left), and the training loss curve (right). The error stays below $5\times10^{-3}$ across most of the parameter range.}
    \label{fig:2dfpeloss}
\end{figure}

The neural network is a U-Net with a single input channel for the scalar parameter $\alpha$ and a single output channel for the density $u$ on a $32 \times 32$ grid, using three down- and up-sampling stages with channel widths $(64,128,256)$ \citep{Ronneberger2015}. Training inputs are normalized constant images built from 256 uniformly sampled values of $\alpha \in [1.0,2.0]$. For each value the network produces a candidate equilibrium density $u_{\mathrm{grid}}(\alpha)$, which is interpreted as a set of cell averages and passed to FiPy as an initial condition, and the solver advances this state for $N_s$ iterations to produce $u_{\mathrm{solver}}(\alpha)$. We train with a batch size of 64 and a learning rate of $10^{-4}$ under cosine annealing \citep{loshchilov2017sgdrstochasticgradientdescent} for 1000 epochs, with the loss converging to the order of $10^{-4}$ as shown in Figure~\ref{fig:2dfpeloss}.

Because the network output serves as a virtual initial condition for the finite volume solver, initialization again plays a central role in training stability. Standard Kaiming initialization produces outputs smooth and bounded enough to ensure solver convergence throughout training \citep{he2015delving}

\begin{figure}[htbp]
    \centering
    \captionsetup{font=small,labelfont=small}
    \includegraphics[width=0.48\textwidth]{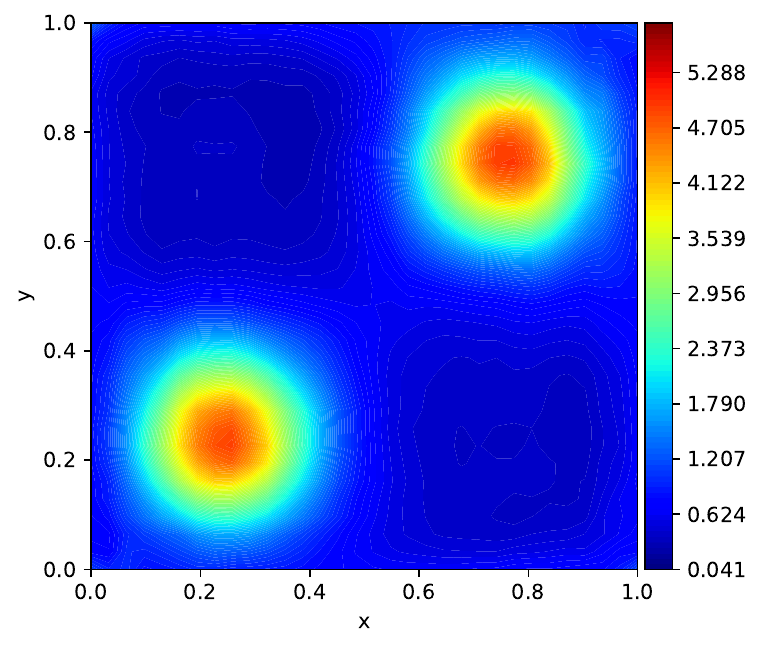}
    \hfill
    \includegraphics[width=0.48\textwidth]{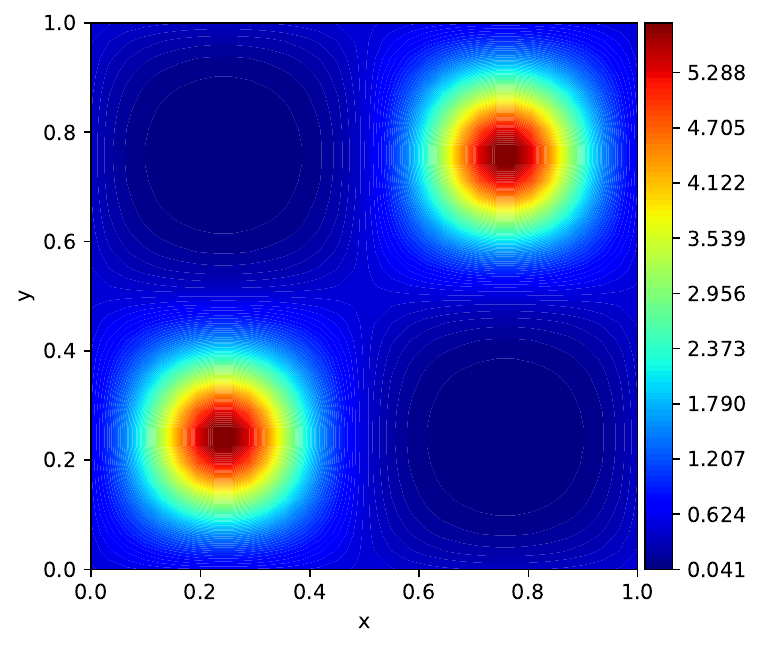}
    \caption{Equilibrium density at $\alpha=1.5$, comparing the \method prediction (left) with the analytical Boltzmann equilibrium (right). The network recovers both the spatial structure of the density and its peak locations.}
    \label{fig:fpe2d}
\end{figure}

As shown in Figure~\ref{fig:fpe2d}, the predicted solution closely matches the analytical equilibrium distribution. Across the full range $\alpha \in [1,2]$ the learned densities track the analytical Boltzmann solutions closely, with the $L^2$ error between the predicted and analytical solutions remaining below $5 \times 10^{-3}$ for most values of $\alpha$, as summarized in Figure~\ref{fig:2dfpeloss}. The network therefore captures both the spatial structure and the amplitude variation induced by changes in $\alpha$, which shows that \method can accurately learn steady-state solutions governed by finite volume discretizations.

\subsection{The Two-dimensional Lid-Driven Cavity using the Finite Element Method (FEM)}
The two-dimensional lid-driven cavity is a classical benchmark in computational fluid dynamics \citep{GHIA1982,Ertuck2005}. A unit square cavity is filled with an incompressible fluid, the top lid moves at a constant unit velocity, and the remaining three walls are stationary. For Reynolds numbers in $\mathrm{Re} \in [2000,3000]$ the flow reaches a steady but strongly nonlinear regime characterized by multiple corner vortices and complex recirculation patterns \citep{ARUN2015,ARUMUGAPERUMAL2011}. Despite the simple geometry, this regime is difficult for data-driven solvers because advection dominates and the velocity gradients are sharp, and standard physics-informed neural networks are known to struggle here at Reynolds numbers of this magnitude \citep{CaoZhang2025}. We therefore apply the \method framework, coupling a U-Net with a high-fidelity finite element solver so that the network acts as a parameter-to-state initializer while the numerical solver enforces the governing equations and boundary conditions.

The lid-driven cavity flow obeys the incompressible Navier--Stokes equations,
\begin{equation}
\label{eq:2dcavity}
\frac{\partial \mathbf{u}}{\partial t}
+ (\mathbf{u}\cdot\nabla)\mathbf{u}
=
-\nabla P
+ \frac{1}{\mathrm{Re}} \nabla^2 \mathbf{u},
\qquad
\nabla\cdot\mathbf{u}=0,
\end{equation}
with $\mathbf{u} = (1,0)$ on the moving lid $\Gamma_0$ and $\mathbf{u} = (0,0)$ on the three stationary walls $\Gamma_1$. The unknowns are the velocity field $\mathbf{u}(\mathbf{x},t) = (u(\mathbf{x},t), v(\mathbf{x},t))$ and the pressure field $P(\mathbf{x},t)$. Our objective is the steady-state equilibrium reached for each $\mathrm{Re} \in [2000,3000]$, which lets us cast the problem as a parameter-to-solution map and motivates the U-Net architecture.

Reference solutions are computed with the finite element method (FEM) in NGSolve \citep{BrennerScottFEM, ngsolve2025}. The domain is discretized by a triangular mesh generated through NGSolve's \texttt{Mesh} routine with a maximum global spacing size of $h_{\max} = 0.05$. Velocity is represented with piecewise cubic basis functions and pressure with piecewise quadratic basis functions. Time integration uses an implicit--explicit (IMEX) scheme with step size $\Delta t = 3\times 10^{-3}$ \citep{LarssonThomeeNumPDE,Boscarino2024}, and for each input configuration the solver runs for $N_s = 500$ iterations. In comparison, for full convergence the solver is need to run for $200,000$ iterations.

The U-Net takes a single input channel encoding the Reynolds number and returns three output channels for the velocity components and pressure on a $32 \times 32$ Cartesian grid, $(\widehat{u}_{\mathrm{grid}}, \widehat{v}_{\mathrm{grid}}, \widehat{P}_{\mathrm{grid}}) = f_{\mathrm{NN}}(\mathrm{Re};\theta)$. It uses three down- and up-sampling stages with channel widths $(64,128,256)$. Training inputs are normalized constant images built from 256 randomly sampled values of $\mathrm{Re} \in [2000,3000]$.

The grid predictions are coupled to the solver through NGSolve's \texttt{VoxelCoefficient} interpolation, which linearly maps the grid values onto the finite element mesh and supplies them as initial conditions. The solver then advances these fields for $N_s = 500$ iterations to produce the converged state $(u^*_{\mathrm{FEM}}, v^*_{\mathrm{FEM}}, P^*_{\mathrm{FEM}})$, which is sampled back onto an evenly spaced $32 \times 32$ lattice to give the targets $(u^*_{\mathrm{grid}}, v^*_{\mathrm{grid}}, P^*_{\mathrm{grid}})$. This formulation decouples the network from a fixed mesh resolution, although meaningful interpolation still requires the grid resolution of the U-Net output to be comparable to the FEM mesh density. We train with the AdamW optimizer at a learning rate of $10^{-5}$ and a batch size of 64 for 300 epochs \cite{AdamWpaper}. As with the Fokker-Planck system previously, we find the default Kaiming initialization (gain $1.0$) in the HuggingFace U-Net implementation sufficient for stable training.

\begin{figure}[h]
    \centering
    \captionsetup{font=small,labelfont=small}
    \includegraphics[width=0.4\textwidth]{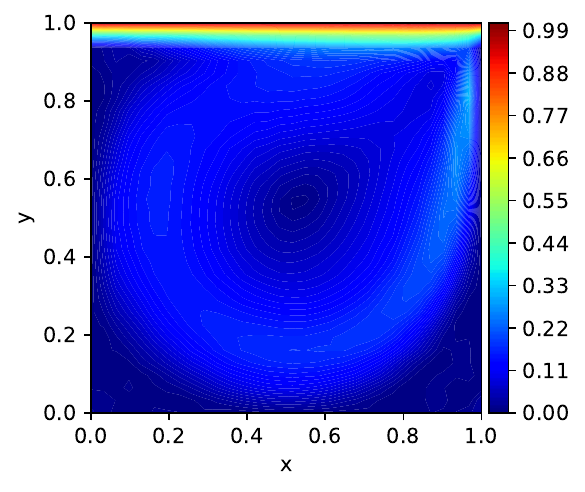}
    \hspace{2mm}
    \includegraphics[width=0.4\textwidth]{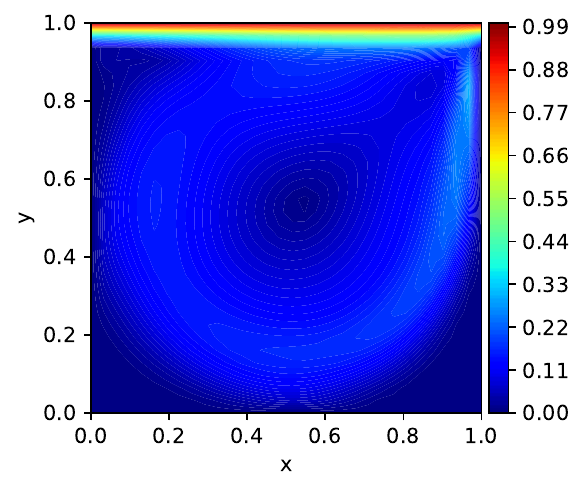}
    \caption{Velocity magnitude at $\mathrm{Re}=2500$, comparing the \method prediction (left) with the FEM reference (right). The network recovers the primary recirculation and the secondary corner vortices.}
    \label{fig:2dcavity_vel}
\end{figure}

\begin{figure}[h]
    \centering
    \captionsetup{font=small,labelfont=small}
    \includegraphics[width=0.4\textwidth]{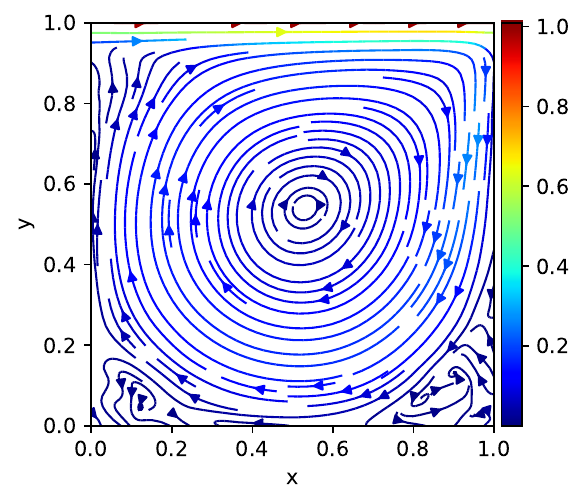}
    \hspace{2mm}
    \includegraphics[width=0.4\textwidth]{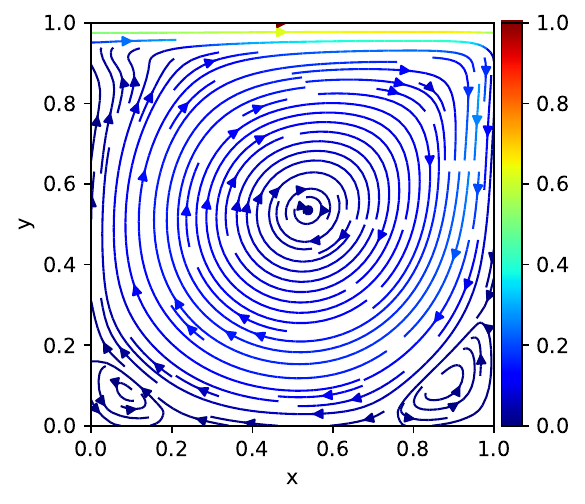}
    \caption{Streamlines at $\mathrm{Re}=2500$ for the \method prediction (left) and the FEM reference (right). The recirculation structure and corner-vortex topology are reproduced.}
    \label{fig:2dcavity_streamlines}
\end{figure}

\begin{figure}[h]
    \centering
    \captionsetup{font=small,labelfont=small}
    \includegraphics[width=0.4\textwidth]{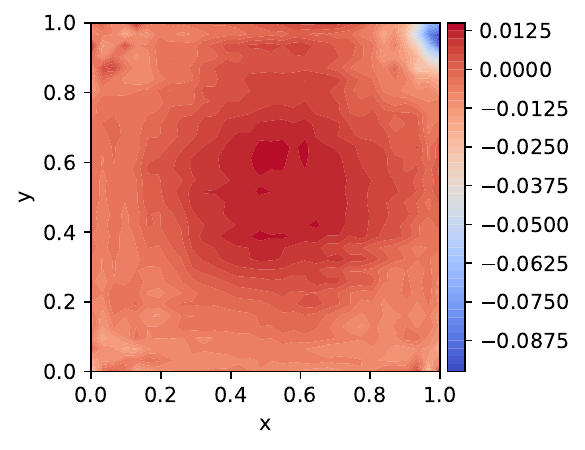}
    \hspace{2mm}
    \includegraphics[width=0.4\textwidth]{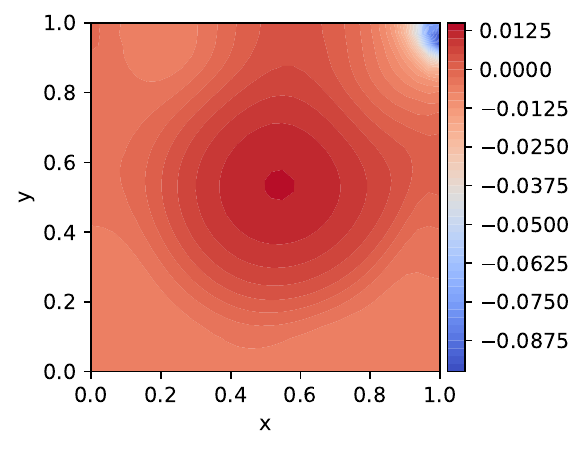}
    \caption{Pressure field at $\mathrm{Re}=2500$ for the \method prediction (left) and the FEM reference (right). The field is near zero over most of the domain, and the model captures the localized negative-pressure region in the upper-right corner.}
    \label{fig:2dcavity_pressure}
\end{figure}

\begin{figure}[h]
    \centering
    \captionsetup{font=small,labelfont=small}
    \includegraphics[width=0.4\textwidth]{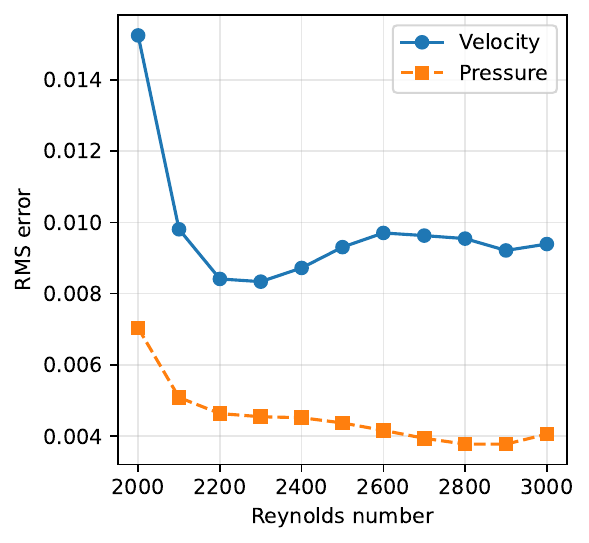}
    \hspace{2mm}
    \includegraphics[width=0.4\textwidth]{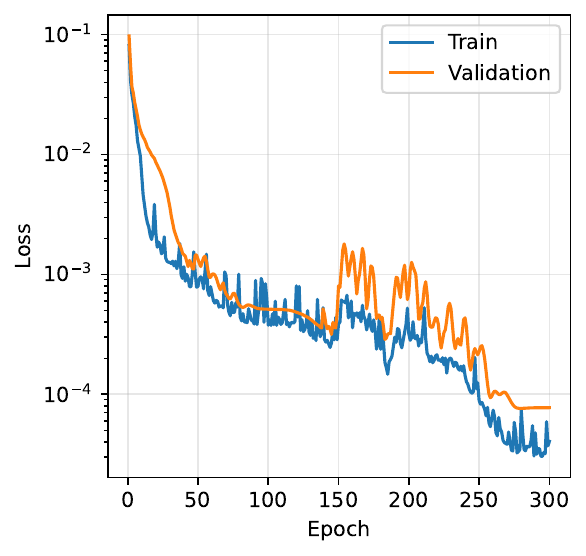}
    \caption{RMS error of the velocity magnitude and pressure as a function of the Reynolds number (left), and the training loss curve (right). Errors stay below $10^{-2}$ across the range and rise near the interval boundaries where the training density is lower.}
    \label{fig:2dcavity_err_loss}
\end{figure}

As shown in Figures~\ref{fig:2dcavity_vel} and~\ref{fig:2dcavity_streamlines}, the predicted velocity field closely matches the reference solution, somewhat capturing both the primary recirculation and the secondary vortex in the bottom-right corner.  Validation losses are computed using mean-squared error across the three physical fields for 64 randomly sampled Reynolds numbers in $[2000,3000]$. Across multiple runs the \method loss typically settles around $10^{-4}$ after roughly 300 epochs, as shown in Figure~\ref{fig:2dcavity_err_loss}. The oscillation in later epochs likely reflects the network refining its predictions in the corner-vortex regions.

The velocity fields agree well in both structure and magnitude, with only minor discrepancies in fine-scale corner vortex features. The pressure fields look visually different largely because they are near zero over most of the domain (and thus have small training gradient values), but the model still captures the localized negative-pressure region in the upper-right corner that governs much of the flow. The RMS errors for velocity magnitude and pressure as functions of the Reynolds number, shown in Figure~\ref{fig:2dcavity_err_loss}, generally remain below $10^{-2}$ across the parameter range. The larger errors near $\mathrm{Re}=2000$ and $\mathrm{Re}=3000$ are likely attributable to the reduced training data density near the interval boundaries. The first-order grid-to-mesh conversion further limits the achievable resolution, since with $h_{\max}=0.05$, a baseline interpolation error on the order of $h_{\max}^2 \approx 2.5\times10^{-3}$ is expected.

\subsection{The One-dimensional Variable-Coefficient Advection Equation using the Finite Volume Method}
We next evaluate the ability of \method to handle transient transport dynamics with parameter-dependent and spatially varying advection speeds. Unlike purely diffusive systems, advection-dominated equations propagate and deform solution profiles along characteristics, which makes them sensitive to both the velocity field and the structure of the initial condition. When the transport speed varies across space the dynamics are no longer simple translations but exhibit local stretching and compression of the profile, so the problem is a useful test of whether \method can learn solver-consistent transient dynamics in a hyperbolic setting.

We consider the one-dimensional variable-coefficient advection equation
\begin{equation}
u_t + v(x;\alpha)\,\frac{\partial u}{\partial x} = 0, \qquad x \in [0,1], \quad t \in [0,0.3],
\end{equation}
with advection velocity 
\begin{equation}
v(x;\alpha) = \alpha + \sin(2\pi x), \qquad \alpha \in [1.5,2.5],
\end{equation} 
where the scalar parameter $\alpha$ sets the mean transport speed and the sinusoidal term introduces spatial variability in the flow. Periodic boundary conditions are imposed on the spatial domain. The initial condition is given by
\begin{equation}
u(x,0) = \exp\left(-50(x-0.3)^2\right) + \mathbbm{1}_{(0.6,0.8)}(x),
\end{equation}
which combines a smooth Gaussian feature with a discontinuous top-hat profile. This mixed initial condition produces a multiscale transient evolution that exercises both smooth transport and sharper advective features.

High-fidelity reference solutions are generated with the CLAWPACK package through its PyClaw interface, using the variable-coefficient one-dimensional advection Riemann solver \citep{clawpack}. The solver is a finite-volume method on a cell-centered grid with $100$ spatial cells over $[0,1]$, with periodic boundary conditions for both the state and the auxiliary velocity field and the MC limiter for total-variation-diminishing reconstruction. The temporal interval uses $\Delta t = 0.05$, and each \method solver call advances the predicted state by one temporal interval, corresponding to a single solver step in the current implementation.

The learning task is to approximate the spatiotemporal solution map $u(x,t;\alpha)$ over the prescribed parameter and domain ranges. We use a fully connected network with inputs $(x,t,\alpha)$, output dimension $1$, hidden width $64$, depth $16$, and SiLU activations. Training data come from $256$ uniformly sampled values of $\alpha \in [1.5,2.5]$, while validation uses $16$ parameter samples together with their exact solution trajectories. The network is trained in a solver-coupled manner with the transient \method loss, in which the prediction at time $t_n$ is passed into the CLAWPACK solver and advanced to $t_{n+1}$, and the solver output is compared against the network prediction at the next time level. An additional initial-condition loss is imposed at $t=0$ with weight $\lambda_{\mathrm{IC}} = 10$. Optimization uses AdamW with learning rate decaying from $5 \times 10^{-4}$ to $5 \times 10^{-5}$ via cosine annealing warm restarts, weight decay $10^{-4}$, batch size $64$, and gradient clipping at $1.0$, over $5000$ epochs. To avoid destabilizing the advection solver in the initial stages of training, the predicted state passed into the solver is clamped to $[-3,3]$. The governing dynamics are then enforced entirely through the \method classical finite-volume solver loss.

\begin{figure}[htbp]
    \centering
    \captionsetup{font=small,labelfont=small}

    \begin{minipage}{0.48\textwidth}
        \centering
        \includegraphics[width=0.83\linewidth]{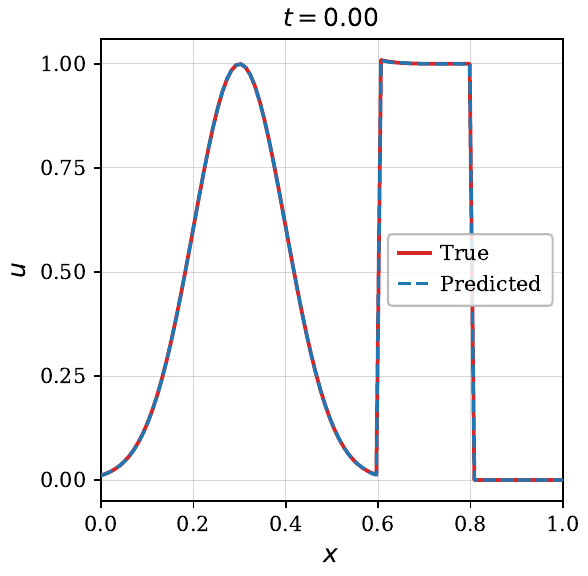}
    \end{minipage}
    \hfill
    \begin{minipage}{0.48\textwidth}
        \centering
        \includegraphics[width=0.83\linewidth]{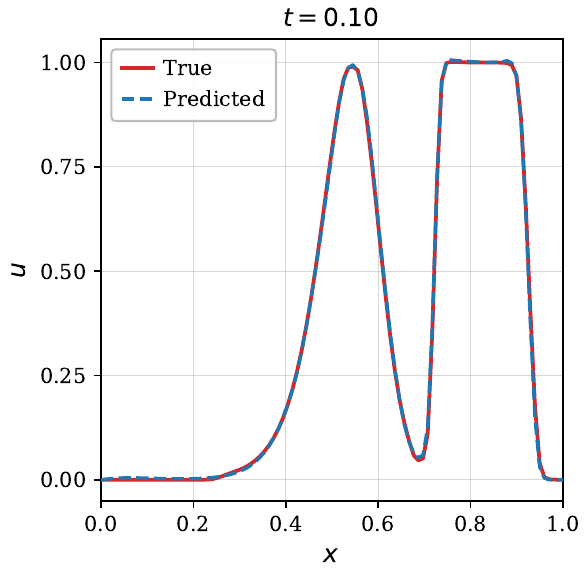}
    \end{minipage}

    \vspace{0.5em}

    \begin{minipage}{0.48\textwidth}
        \centering
        \includegraphics[width=0.83\linewidth]{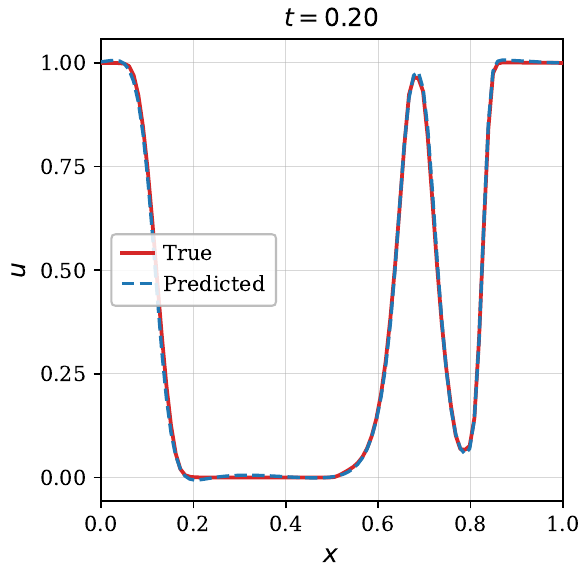}
    \end{minipage}
    \hfill
    \begin{minipage}{0.48\textwidth}
        \centering
        \includegraphics[width=0.83\linewidth]{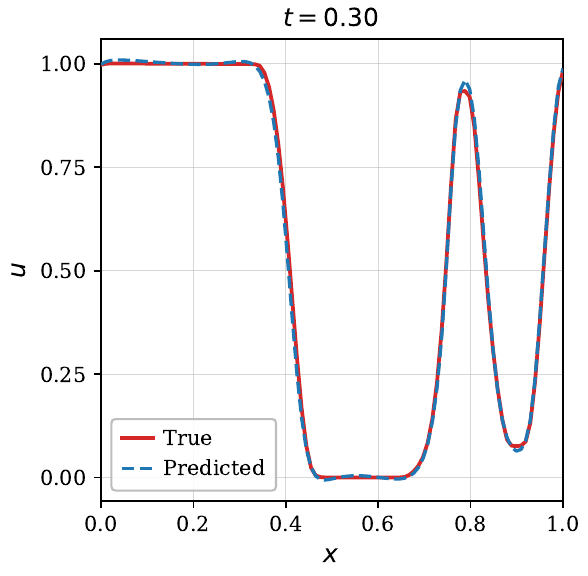}
    \end{minipage}

    \caption{Snapshots of the 1D advection solution at $\alpha = 2.0$ for $t=0$, $0.10$, $0.20$, and $0.30$, comparing the \method prediction with the PyClaw reference. The model tracks both the Gaussian bump and the top-hat discontinuity as they advect and deform.}
    \label{fig:ad1d}
\end{figure}

As shown in Figure~\ref{fig:ad1d}, the predicted solution matches the reference numerical solution closely and accurately captures the initial discontinuities and their spreading, with the RMS error generally being under 0.01 for most parameter values. 

\begin{figure}[htbp]
    \centering
    \captionsetup{font=small,labelfont=small}

    \begin{minipage}{0.48\textwidth}
        \centering
        \includegraphics[width=0.83\linewidth]{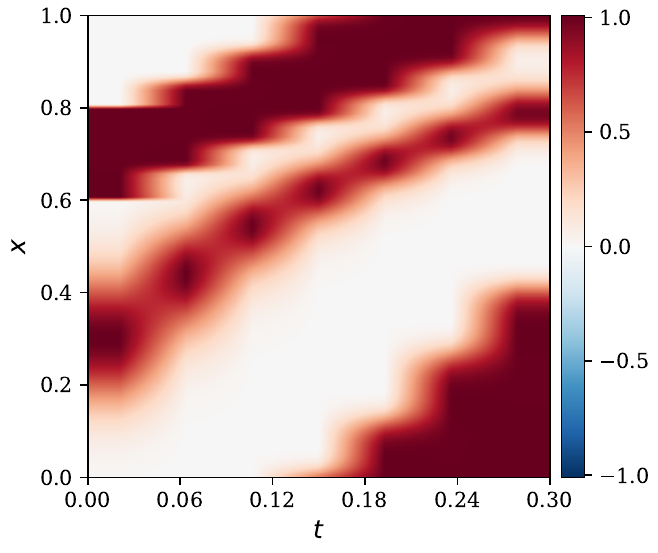}
    \end{minipage}
    \hfill
    \begin{minipage}{0.48\textwidth}
        \centering
        \includegraphics[width=0.83\linewidth]{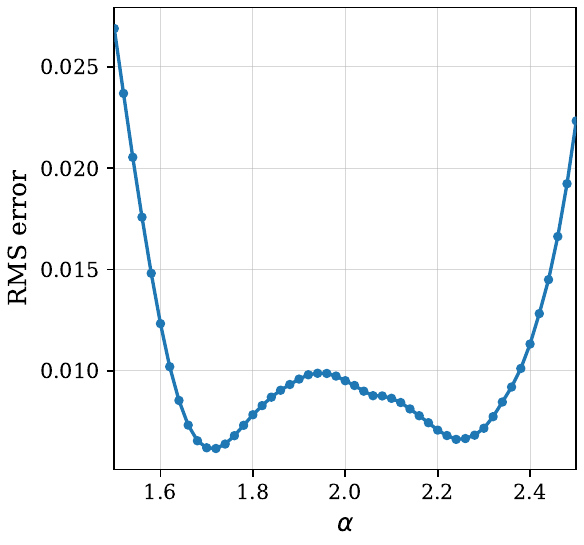}
    \end{minipage}

    \caption{Space-time solution trajectory predicted by \method at $\alpha = 2.0$ (left), and the RMS error against the reference solution as a function of $\alpha$ (right).}
    \label{fig:ad1d_error}
\end{figure}

We next consider the corresponding inverse problem for the same one-dimensional variable-coefficient advection equation. Rather than learning the solution map over a family of parameters, the goal here is to recover the unknown advection parameter from noisy observations of the transported state. Since the advection speed directly controls the location and deformation of the propagated features, this provides a natural test of the robustness of \method in a solver-coupled inverse setting.

Synthetic inverse data are generated by evolving the system with the ground-truth parameter $\alpha^\ast=2.0$ up to the final observation time $t_{\mathrm{obs}}=0.3$ and then perturbing the resulting solution with additive Gaussian noise, and we report results for relative noise levels of $5\%$, $10\%$, and $15\%$.

The architecture is identical to the forward setting, with the key difference that the unknown parameter is now treated as a trainable scalar optimized jointly with the network. The network remains constrained through the transient \method loss, while the inverse parameter is updated by matching the predicted solution at the observation time to the noisy measurements. Thus the unknown transport coefficient is recovered through the same solver-coupled framework used in the forward problem without any explicit PDE residual terms. Because errors in the advection parameter accumulate over time and lead to a global transport mismatch, the $5\%$, $10\%$, and $15\%$ noise cases provide a useful measure of the robustness of the inverse formulation under increasing observational uncertainty.

Table~\ref{tab:ad1d_inverse} reports the recovered advection parameter at each noise level, where the optimization is initialized at $\alpha = 1.5$ on the lower edge of the search interval $[1.5,2.5]$ so that the procedure must traverse the full range rather than remain near the ground truth. The values are averaged over three independent runs, with the reported uncertainty corresponding to the standard deviation across runs. At the lowest noise level the recovered parameter is $2.0017 \pm 0.0023$, and even under $15\%$ relative noise the estimate remains $2.0080 \pm 0.0028$, an absolute error below $0.5\%$ of the true value $\alpha^\ast = 2.0$. The mean absolute error grows monotonically with the noise level, from $1.7 \times 10^{-3}$ at $5\%$ to $8.0 \times 10^{-3}$ at $15\%$, indicating a graceful degradation of accuracy rather than a breakdown of the identification.

These results show that the solver-coupled inverse formulation recovers the transport coefficient accurately and remains stable under substantial observational noise. Since the gradients for the unknown parameter flow only through the neural surrogate while the finite-volume solver enforces the forward dynamics, the quality of the recovery reflects the smoothness of the learned parameter dependence established during the forward training. This behavior is consistent with the inverse results obtained for the other benchmark systems considered here and confirms that \method extends to inverse problems in the hyperbolic transport setting.

\begin{table}[t]
    \centering
    \captionsetup{font=small,labelfont=small}
    \caption{Recovered advection parameter $\alpha$ for the inverse variable-coefficient advection problem under increasing observational noise. The ground-truth value is $\alpha^\ast = 2.0$ and the optimization is initialized at $\alpha = 1.5$. Reported values are the mean over three independent runs with the standard deviation across runs, and the final column lists the absolute error of the mean.}
    \label{tab:ad1d_inverse}
    \begin{tabular}{lcc}
        \toprule
        Noise level & Recovered $\alpha$ & Absolute error \\
        \midrule
        $5\%$  & $2.0017 \pm 0.0023$ & $1.7 \times 10^{-3}$ \\
        $10\%$ & $2.0028 \pm 0.0125$ & $2.8 \times 10^{-3}$ \\
        $15\%$ & $2.0080 \pm 0.0028$ & $8.0 \times 10^{-3}$ \\
        \bottomrule
    \end{tabular}
\end{table}


\subsection{The Two-dimensional Transient Flow Past a Cylinder using the Finite Element Method}

We now move onto higher dimension problems with more complex spatial geometries. We consider two-dimensional incompressible flow past a circular cylinder confined in a channel. This is a canonical wind tunnel benchmark in computational fluid dynamics. For sufficiently high Reynolds numbers, the wake behind the cylinder becomes unstable and develops a periodic von K\'arm\'an vortex street \citep{Albarede_Provansal_1995, Monkewitz1996}. This problem tests the ability of the \method framework to handle complex boundary conditions and transient dynamics.

The flow is governed by the incompressible Navier-Stokes equations,
\begin{equation}
\frac{\partial \mathbf{u}}{\partial t} + (\mathbf{u}\cdot\nabla)\mathbf{u}
= -\nabla P + \frac{1}{Re}\nabla^2 \mathbf{u},
\qquad
\nabla \cdot \mathbf{u} = 0,
\label{eq:wt-ns}
\end{equation}
where $\mathbf{u}(\mathbf{x},t) = (u(\mathbf{x},t), v(\mathbf{x},t))$ is the velocity field and $P(\mathbf{x},t)$ is the pressure. The spatial domain is the channel $\Omega = [0, 2.0] \times [0, 0.41]$ with a circular obstacle of radius $r = 0.05$ centered at $(0.2, 0.2)$, and the slight vertical offset of the cylinder relative to the channel centerline breaks the symmetry of the wake and promotes the onset of shedding. A parabolic velocity profile $\mathbf{u}_{\mathrm{in}}(y) = \left( u_{\max}\,4\,y\,(L_y - y)/L_y^2,\; 0 \right)$ with $u_{\max} = 1.5$ is prescribed at the left inlet. A natural do-nothing condition is imposed at the right outlet, and no-slip conditions are imposed on the top and bottom walls and on the cylinder surface. We define the Reynolds number through the mean inlet velocity and the cylinder diameter as $Re = U_{\mathrm{mean}}\, D / \nu$ with $U_{\mathrm{mean}} = \tfrac{2}{3}\,u_{\max}$ and $D = 2r$. We study the parameter range $Re \in [50, 100]$, which corresponds to kinematic viscosities $\nu \in [10^{-3}, 2\times 10^{-3}]$. This interval spans the transition from a nearly steady wake near the shedding onset to a clearly periodic vortex street at the upper end, so a single network is asked to generalize across qualitatively different flow regimes.

Reference dynamics are produced by a finite element solver implemented in NGSolve \citep{ngsolve2025}. The channel with the cylindrical cutout is meshed with triangular elements at a maximum mesh spacing of $h_{\max} = 0.07$, and the mesh is curved to third order to represent the cylinder boundary accurately. The velocity is discretized with piecewise cubic basis functions and the pressure with piecewise quadratic basis functions. Temporal advancement uses a linearly implicit-explicit scheme in which the viscous and pressure coupling is treated implicitly through a Stokes operator $A$ while the nonlinear convection $C(\mathbf{u})$ is treated explicitly. With mass matrix $M$ and pseudo-time step $\Delta t = 10^{-3}$. The operator $M + \Delta t\,A$ is factorized once by a sparse Cholesky decomposition and reused, and each substep applies the update
\begin{equation}
\mathbf{u} \;\leftarrow\; \mathbf{u} \;-\; \Delta t\,\bigl(M + \Delta t\,A\bigr)^{-1}\bigl(C(\mathbf{u}) + A\,\mathbf{u}\bigr).
\label{eq:wt-substep}
\end{equation}
Each network time step advances the flow by $\Delta t_{\mathrm{net}} = 0.1$, corresponding to $N_s = 100$ solver substeps of the update in \eqref{eq:wt-substep}, and we evolve the system over $t \in [0, 1.0]$ for $N_t = 10$ network time steps. The continuous fields are sampled onto a uniform $256 \times 64$ grid for the network, and a binary geometry mask is recorded so that all quantities inside the cylinder are set to zero.

The neural network is a U-Net that maps the Reynolds number and the channel geometry to the full flow state, taking two spatial input channels, namely a constant image of the normalized Reynolds number and the binary geometry mask, and producing three output channels for the velocity components and the pressure. It uses four down- and up-sampling stages with channel widths $(64, 128, 256, 512)$, two residual layers per stage, group normalization, and SiLU activations, and its output is multiplied by the geometry mask so that the cylinder interior is enforced to be zero.

One key architectural choice for this transient problem concerns how time enters the network. We inject the normalized time $t_n$ through the sinusoidal timestep embedding native to the U-Net rather than as a spatial channel. This embedding expands a scalar time into a multi-frequency representation, which provides the network with the high-frequency information needed to distinguish phases of the shedding cycle and to represent the oscillatory dynamics.

Training inputs consist of $256$ Reynolds numbers sampled uniformly from $[50, 100]$, and validation uses $16$ Reynolds numbers spaced evenly across the same interval. The model is trained with the AdamW optimizer at a learning rate of $10^{-5}$, a weight decay of $10^{-4}$, and a batch size of $8$ for up to $2000$ epochs. A monotone cosine annealing schedule is applied to the learning rate to avoid loss spikes.

\begin{figure}[h]
  \centering
  \setlength{\tabcolsep}{1pt}
  \renewcommand{\arraystretch}{0.4}
  \begin{tabular}{c cc}
     & Prediction & Reference \\
    \rotatebox[origin=c]{90} &
      \includegraphics[width=0.42\textwidth]{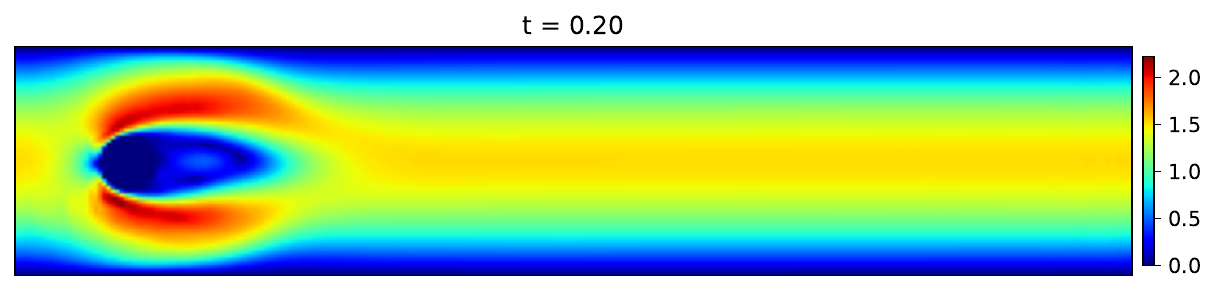} &
      \includegraphics[width=0.42\textwidth]{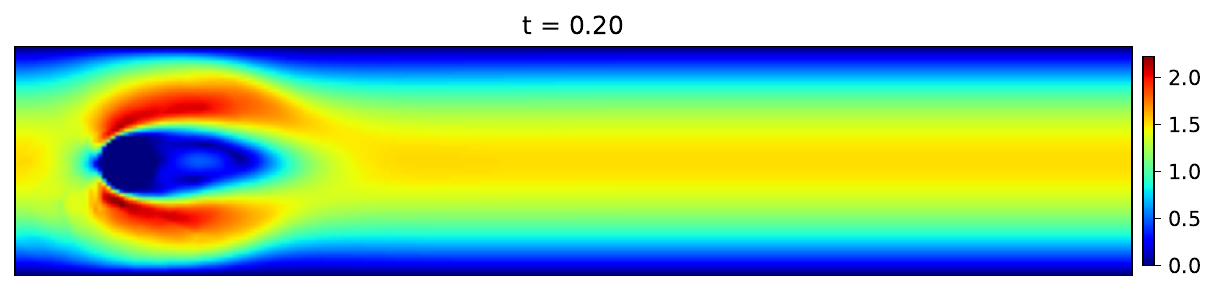} \\
    \rotatebox[origin=c]{90} &
      \includegraphics[width=0.42\textwidth]{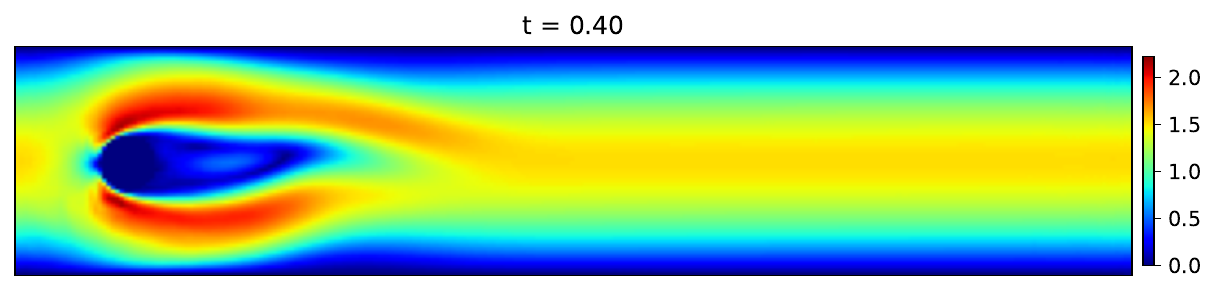} &
      \includegraphics[width=0.42\textwidth]{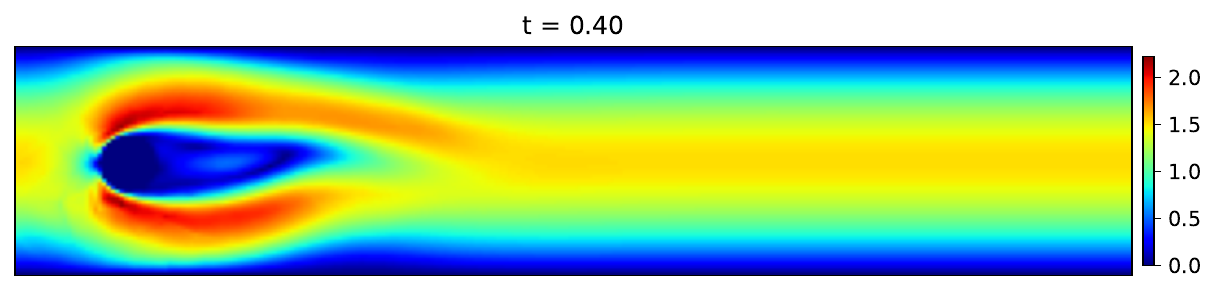} \\
    \rotatebox[origin=c]{90} &
      \includegraphics[width=0.42\textwidth]{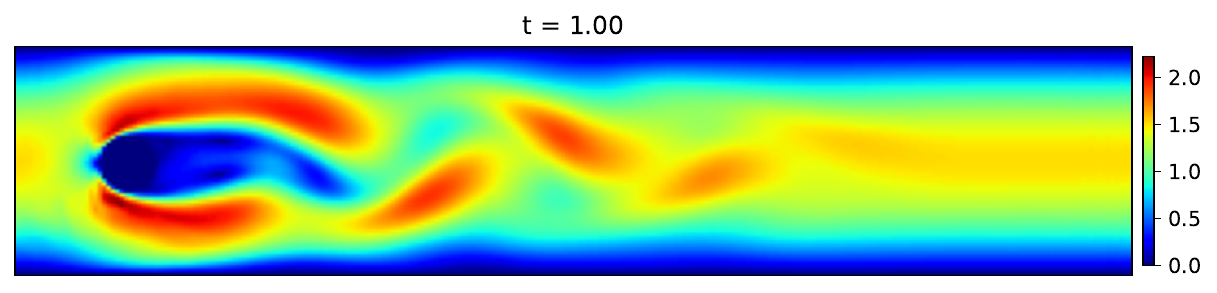} &
      \includegraphics[width=0.42\textwidth]{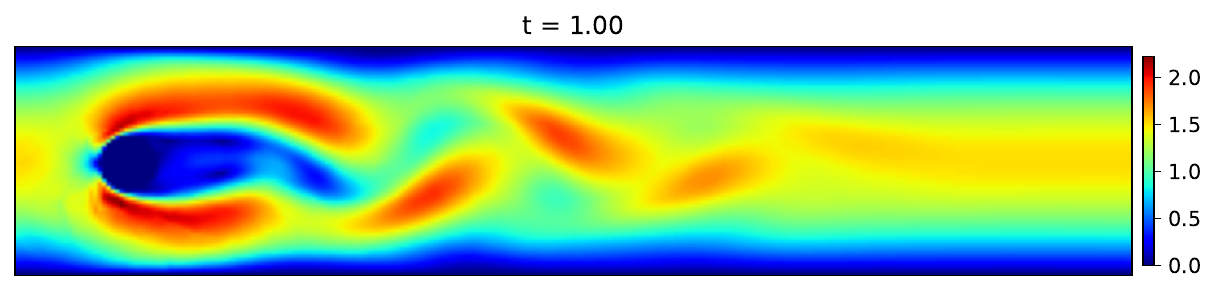} \\
  \end{tabular}
  \caption{Predicted and reference velocity magnitude for flow past the cylinder
  at $Re=75$ at three time instants. The left column is the \method prediction
  and the right column is the finite element reference, and the prediction tracks
  the developing wake and the alternating structure of the von K\'arm\'an vortex street.}
  \label{fig:wt-velocity}
\end{figure}

\begin{figure}[h]
  \centering
  \setlength{\tabcolsep}{1pt}
  \renewcommand{\arraystretch}{0.4}
  \begin{tabular}{c cc}
     & Prediction & Reference \\
    \rotatebox[origin=c]{90} &
      \includegraphics[width=0.42\textwidth]{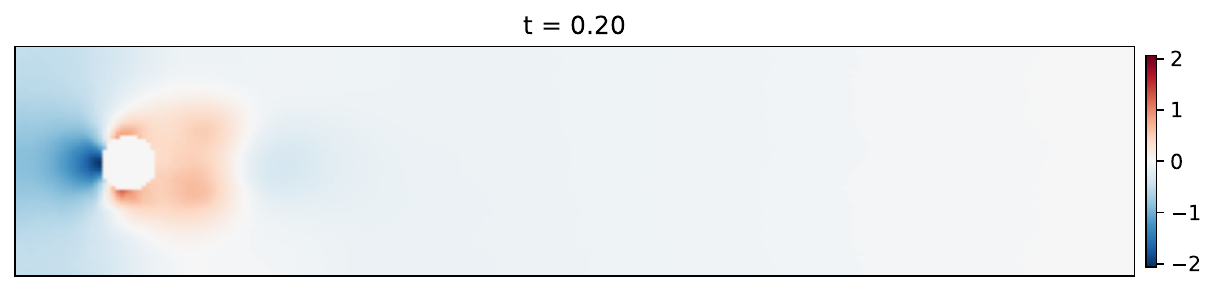} &
      \includegraphics[width=0.42\textwidth]{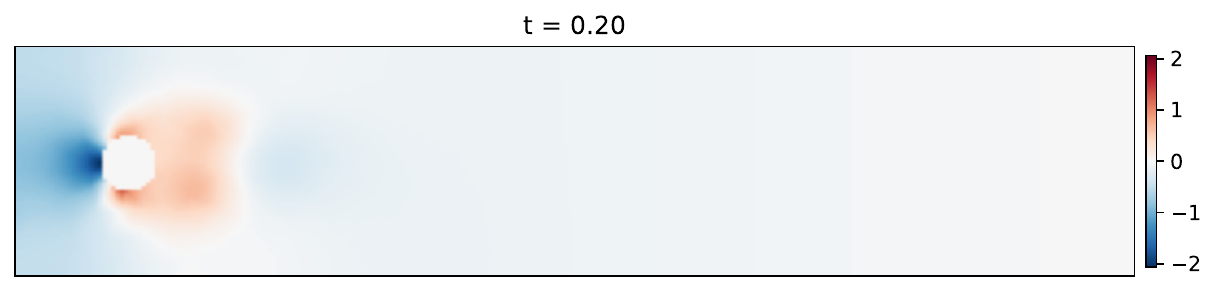} \\
    \rotatebox[origin=c]{90} &
      \includegraphics[width=0.42\textwidth]{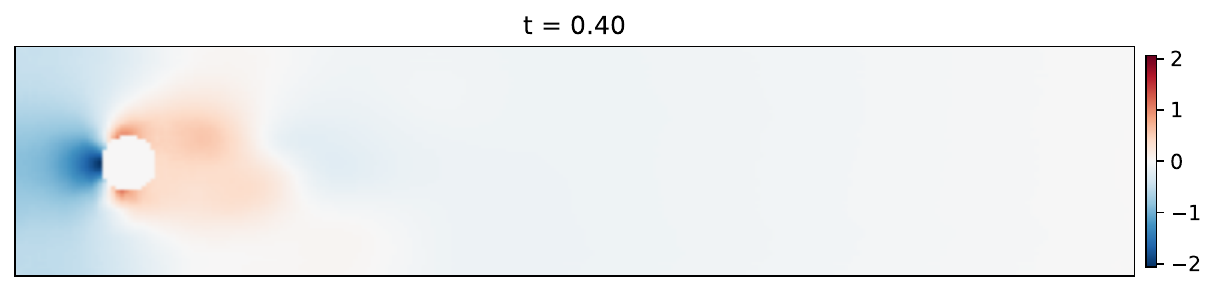} &
      \includegraphics[width=0.42\textwidth]{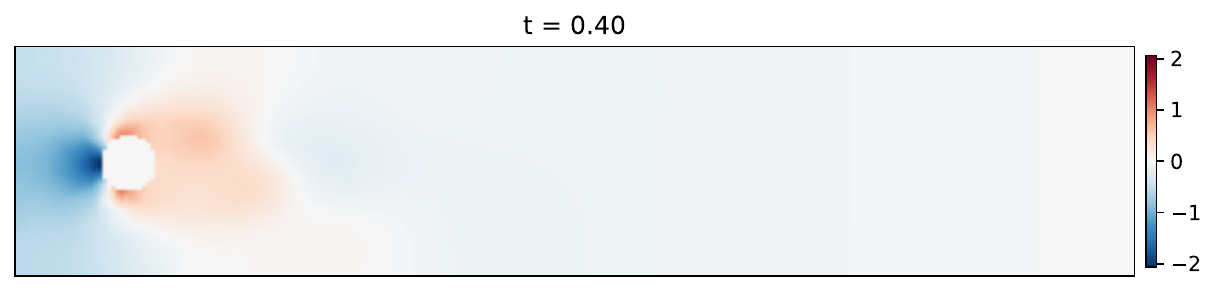} \\
    \rotatebox[origin=c]{90} &
      \includegraphics[width=0.42\textwidth]{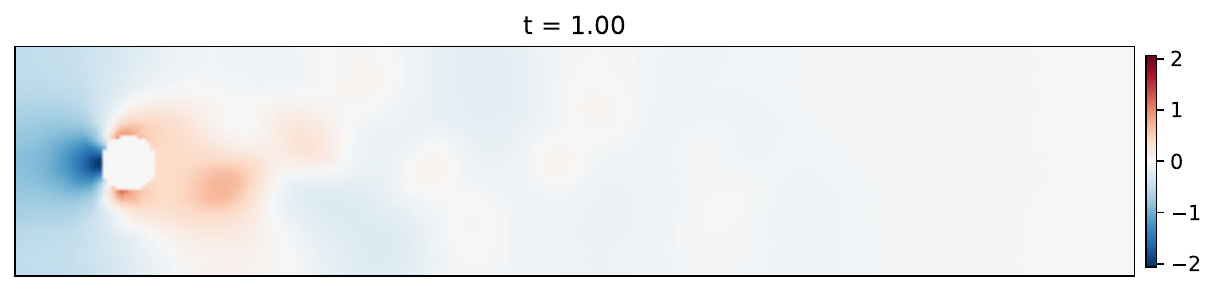} &
      \includegraphics[width=0.42\textwidth]{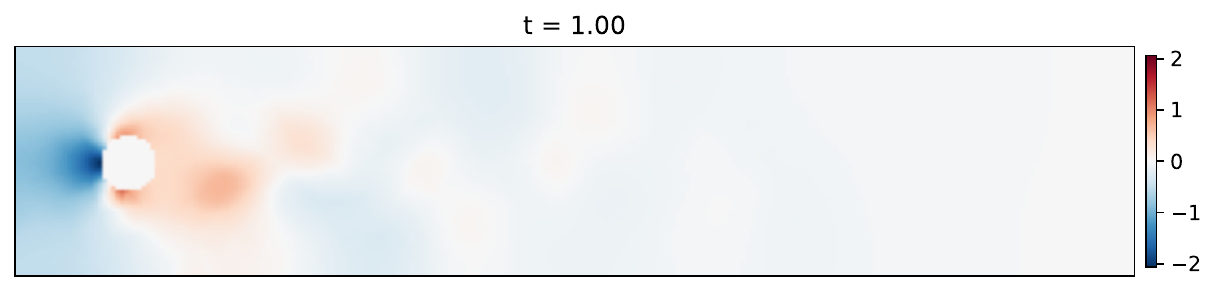} \\
  \end{tabular}
  \caption{Predicted and reference pressure field for flow past the cylinder at
  $Re=75$ at three time instants. The left column is the \method prediction and
  the right column is the finite element reference, and the prediction reproduces
  the pressure structure around the cylinder and through the wake.}
  \label{fig:wt-pressure}
\end{figure}

Figure~\ref{fig:wt-velocity} and Figure~\ref{fig:wt-pressure} compare the predicted and reference velocity magnitude and pressure fields at three time instants for a representative Reynolds number of $75$. The predicted velocity field reproduces the developing wake behind the cylinder and the alternating structure of the von K\'arm\'an vortex street, and the predicted pressure field recovers the high pressure region at the front of the cylinder together with the low pressure pockets that form and convect downstream in the wake. This confirms that the sinusoidal time embedding supplies the phase information needed to represent the oscillatory regime.

\begin{figure}[h]
  \centering
  \includegraphics[width=0.5\textwidth]{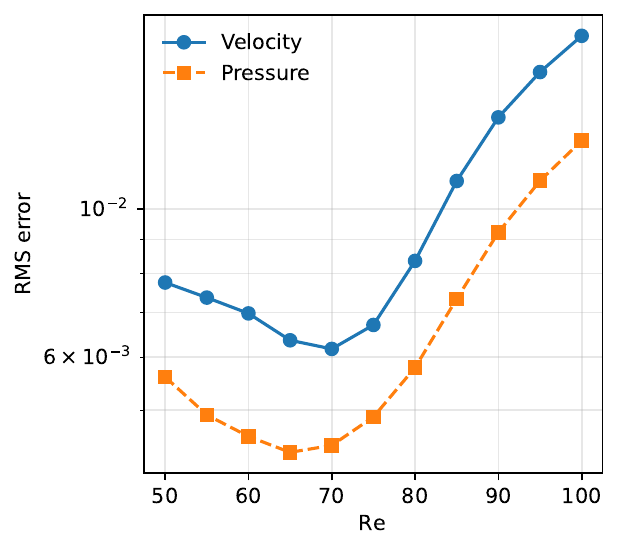}
  \caption{Root mean squared error of the predicted velocity magnitude and
  pressure against the finite element reference as a function of the Reynolds
  number, evaluated over the full transient. The error is smallest in the lower
  and central part of the range and grows toward higher Reynolds numbers as the
  wake becomes more strongly oscillatory.}
  \label{fig:wt-rms}
\end{figure}

We quantify accuracy through the root mean squared error of the velocity magnitude and the pressure against the finite element reference, evaluated over the full transient and reported as a function of the Reynolds number in Figure~\ref{fig:wt-rms}. For the representative case at $Re = 75$ the root mean squared errors of the velocity magnitude and the pressure are $6.7\times 10^{-3}$ and $4.9\times 10^{-3}$ respectively. Averaged across the eleven evaluation Reynolds numbers the errors are $9.9\times 10^{-3}$ for the velocity magnitude and $6.8\times 10^{-3}$ for the pressure, which is well below the characteristic scale of the fields and shows that the surrogate stays faithful to the solver across the parameter range. The error is smallest in the lower and central part of the interval, where the velocity magnitude error falls to about $6.2\times 10^{-3}$ near $Re = 70$, and it rises toward the upper end of the range to $1.8\times 10^{-2}$ for the velocity magnitude and $1.3\times 10^{-2}$ for the pressure at $Re = 100$. This trend follows the underlying physics, because the wake becomes increasingly unsteady and the vortex street more vigorous as the Reynolds number grows, so the high $Re$ regime presents sharper spatial gradients and stronger oscillations that are harder to reproduce. A milder increase is also visible at the lower boundary of the interval, where the network has less neighboring training support to draw on. These results show that the sinusoidal time embedding allows the network to capture oscillatory wake dynamics under solver-consistency training, extending \method from steady-state flow to genuinely transient and periodic regimes.

We turn from the forward surrogate to the inverse problem of recovering an unknown Reynolds number from noisy observation. We generate a single observation snapshot of the velocity and pressure fields at $t = 1.0$ by rolling out the finite element solver at a true Reynolds number of $75$, corrupt it with zero-mean Gaussian noise whose standard deviation is a fixed fraction of the peak field magnitude, and then jointly optimize the network weights against the forward solver-consistency objective and a single scalar Reynolds number against the noisy observation. The scalar is initialized at a deliberately distant guess of $\text{Re} = 60$ and clamped to the search interval $[50, 100]$. We repeat the experiment at three noise levels of $5\%$, $10\%$, and $15\%$, running three independent trials at each level to gauge the run-to-run variability of the recovered value.

Reported as a mean and a standard deviation over the three trials at each level, the recovered Reynolds numbers are $74.94 \pm 0.23$ at $5\%$ noise, $75.12 \pm 0.10$ at $10\%$, and $74.93 \pm 0.06$ at $15\%$, and Table~\ref{tab:wt-inverse} collects these values alongside the relative error of each mean. At every noise level the mean estimate lies within about $0.12$ of the true value of $75$, so the relative error of the mean never exceeds about $0.16\%$, and averaging across all nine runs returns a recovered Reynolds number of $75.00$. The estimates scatter symmetrically about the truth rather than drifting in one direction, which indicates that the recovery is essentially unbiased.

\begin{table}[h]
  \centering
  \caption{Reynolds number recovered from a single noisy observation snapshot
  at a true value of $Re = 75$, with the scalar initialized at $Re = 60$. Each
  row reports the mean and standard deviation over three independent trials at
  the given noise level together with the relative error of the mean. The
  estimate stays accurate and its spread remains small as the noise increases.}
  \label{tab:wt-inverse}
  \begin{tabular}{ccc}
    \toprule
    Noise level & Recovered $Re$ & Relative error \\
    \midrule
    $5\%$  & $74.94 \pm 0.23$ & $0.08\%$ \\
    $10\%$ & $75.12 \pm 0.10$ & $0.16\%$ \\
    $15\%$ & $74.93 \pm 0.06$ & $0.09\%$ \\
    \bottomrule
  \end{tabular}
\end{table}

One feature of note is that the spread of the recovered value stays small at every level and does not grow as the observation becomes noisier, with standard deviations of $0.23$, $0.10$, and $0.06$ at the three noise levels and a mean estimate whose distance from the truth shows the same insensitivity to the noise. This robustness follows from the structure of the inverse objective, which compares the predicted field against the noisy snapshot over the entire observation domain rather than at any one location. Because the per-pixel noise is zero mean and the recovered parameter is determined by the global shape of the velocity and pressure fields, the many spatial samples average the corruption away and the estimate is governed by the field structure rather than by individual noisy values. The dominant source of the small residual offset is consequently not the observation noise but the limited sensitivity of the field to fine changes in the Reynolds number near the optimum together with the tolerance of the scalar optimization. Taken together these results show that solver-consistency training supports accurate and noise-robust parameter recovery on a transient periodic flow, recovering the Reynolds number to within a fraction of a percent from a single corrupted snapshot even when that snapshot carries fifteen percent noise.

\subsection{The Two-dimensional Shallow Water Equations using the Finite Volume Method}

We next apply the \method framework to a nonlinear hyperbolic system of conservation laws. The two dimensional shallow water equations describe the depth-averaged motion of a thin layer of fluid under the force of gravity. The radial dam break configuration is a standard benchmark in which a circular column of elevated water is released and collapses outward. This problem is demanding for two reasons. First, the discontinuous initial data produces an outward-moving circular bore together with an inward-moving rarefaction, so the network must represent sharp features rather than smooth profiles. Second, the parameter of interest controls the strength of the initial imbalance and therefore the speed and amplitude of the resulting shock, so the network must learn an entire family of discontinuous solutions rather than a single trajectory. The finite volume solver provides the consistency operator, and the learnable map is conditioned on the inner dam height.

\begin{figure}[h]
\centering
\newcommand{\swewidth}{0.30\linewidth}
\setlength{\tabcolsep}{2pt}
\renewcommand{\arraystretch}{0.3}
\begin{tabular}{cc}
\footnotesize \method prediction & \footnotesize Finite volume reference  \\
\includegraphics[width=\swewidth]{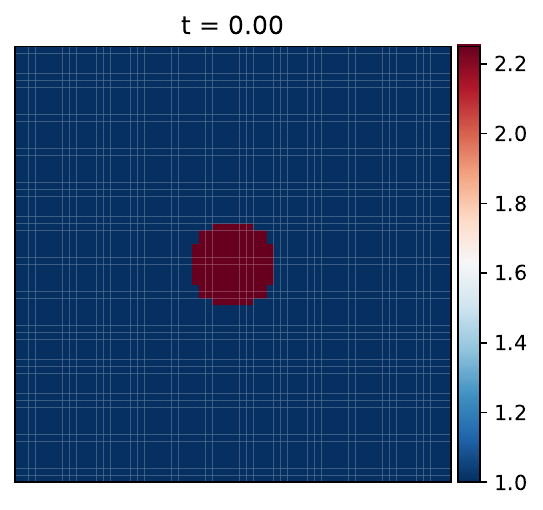} &
\includegraphics[width=\swewidth]{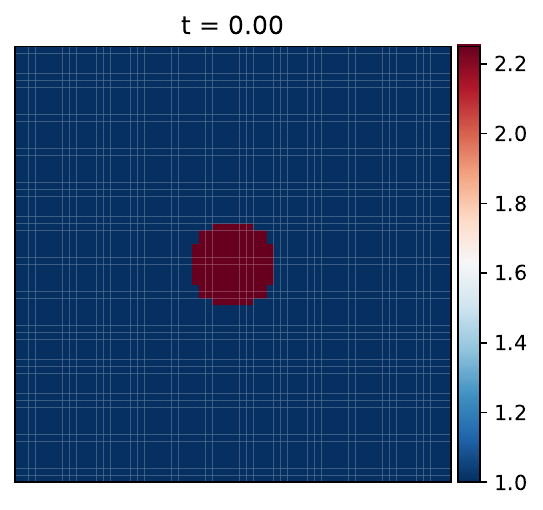} \\
\includegraphics[width=\swewidth]{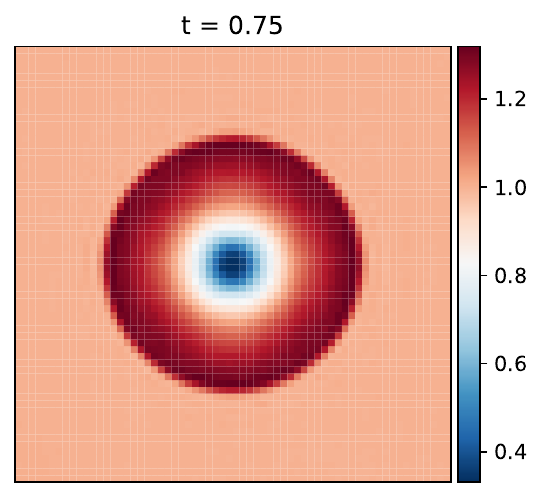} &
\includegraphics[width=\swewidth]{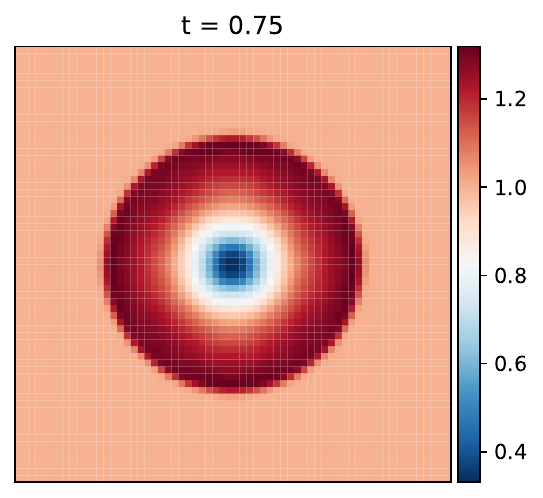} \\
\includegraphics[width=\swewidth]{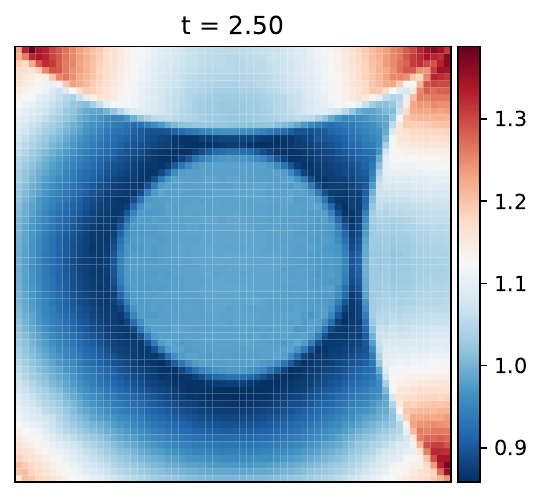} &
\includegraphics[width=\swewidth]{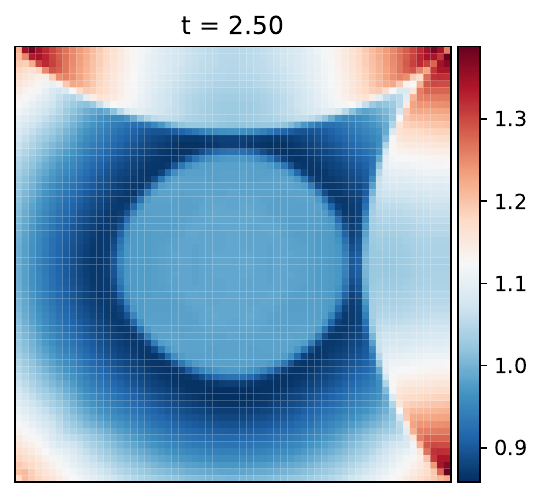} \\
\end{tabular}
\caption{Predicted water depth for the radial dam break at inner dam height $h_{\mathrm{in}} = 2.25$, with the finite volume reference in the left column and the network prediction in the right column and time increasing from top to bottom. The outward circular bore and the inner rarefaction are reproduced across the time horizon, and the prediction stays close to the reference even as the bore sharpens and moves outward.}
\label{fig-swe-fields}
\end{figure}

We consider the two dimensional shallow water equations written in conservation form,
\begin{align}
\frac{\partial h}{\partial t} + \frac{\partial (hu)}{\partial x} + \frac{\partial (hv)}{\partial y} &= 0, \label{eq-swe-mass}\\[2pt]
\frac{\partial (hu)}{\partial t} + \frac{\partial}{\partial x}\!\left(hu^2 + \tfrac{1}{2} g h^2\right) + \frac{\partial (huv)}{\partial y} &= 0, \label{eq-swe-momx}\\[2pt]
\frac{\partial (hv)}{\partial t} + \frac{\partial (huv)}{\partial x} + \frac{\partial}{\partial y}\!\left(hv^2 + \tfrac{1}{2} g h^2\right) &= 0, \label{eq-swe-momy}
\end{align}
where $h$ denotes the water depth, $(hu, hv)$ are the momentum components, and $g$ is the gravitational acceleration which is fixed at $g = 1$. The spatial domain is $\Omega = [-2.5, 2.5]^2$ and the time horizon is $[0, 2.5]$. The initial condition is a radial dam break in which the depth is set to $h_{\mathrm{in}}$ inside a disk of radius $r_d = 0.5$ and to the outer depth $h_{\mathrm{out}} = 1$ outside it, with zero initial momentum, so that $h(\mathbf{x},0) = h_{\mathrm{in}}$ for $\lVert \mathbf{x} \rVert \le r_d$ and $h(\mathbf{x},0) = h_{\mathrm{out}}$ otherwise. The map is conditioned on the inner dam height $h_{\mathrm{in}} \in [1.5, 3.0]$, which sets the size of the initial pressure imbalance and hence the strength of the emerging shock. Here we treat $h_{\mathrm{in}}$ as the physical parameter $\alpha$ in the transient formulation of \method.

Reference solutions are produced with the finite volume method implemented in PyClaw, which is part of the Clawpack family of wave-propagation solvers \citep{clawpack}. The domain is discretized on a uniform Cartesian grid of $64 \times 64$ cells, giving a spatial resolution of $\Delta x = \Delta y = 5/64$. Interface fluxes are computed with a Roe approximate Riemann solver augmented by an entropy fix with a monotonized central total variation diminishing limiter is applied so that spurious oscillations near the bore are suppressed. The two spatial sweeps are combined through dimensional splitting. Outflow conditions are imposed on the lower $x$ and $y$ boundaries through zero-order extrapolation, while solid wall conditions are imposed on the upper $x$ and $y$ boundaries. Each solver call advances the state by one macro time step of size $\Delta t = 0.25$, and the number of internal sub-steps $N_s$ is selected adaptively so that the CFL stability condition is respected, so the time horizon is represented by the eleven points $t \in \{0.0, 0.25, 0.5, \ldots, 2.5\}$.

We use a U-Net architecture in which the two scalar inputs are broadcast into constant-valued images, so that the input tensor carries two channels holding the time and the parameter and the output tensor carries three channels holding the depth and the two momentum components. The network uses three down- and up-sampling stages with channel widths $(32, 64, 128)$, two residual layers per stage, group normalization, and SiLU activations, and its weights are set with Kaiming initialization. The regularizer weight $\lambda_{IC}$ of the initial condition loss is set to 10.

Training proceeds in two phases. In the first phase, the neural network is warm-started for $10$ epochs by regressing directly onto precomputed solver trajectories for a small set of $16$ parameter values, which places the network output inside the region of admissible states for the finite volume scheme. In the second phase the network is trained with the total solver-consistency objective over a larger set of $256$ parameter values. Since the finite volume scheme requires a positive water depth, the predicted depth is clamped to a small positive floor before each solver call. The warm-start, training, and validation parameter sets are randomly drawn from $h_{\mathrm{in}} \in [1.5, 3.0]$, so that evaluation is performed only on inner dam heights that were never seen during training.

\begin{figure}[h]
\centering
\includegraphics[width=0.6\linewidth]{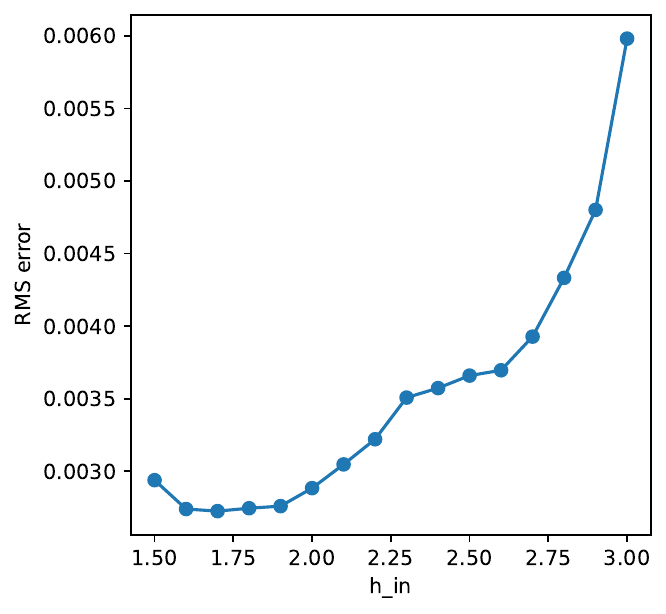}
\caption{RMS error of the predicted state, averaged over the time horizon, as a function of the inner dam height $h_{\mathrm{in}}$. The error grows with $h_{\mathrm{in}}$ as larger initial imbalances produce stronger shocks, reaching about $6.0 \times 10^{-3}$ at the upper end of the range.}
\label{fig-swe-error}
\end{figure}

Optimization uses AdamW with a weight decay of $10^{-4}$ and gradient clipping. The learning rate is $10^{-3}$ during the warm-start phase and $10^{-4}$ during the solver-consistency phase, and within each phase it follows a cosine annealing schedule. We use a batch size of 8.

We evaluate the trained network by comparing its predicted trajectories against independent solver references generated for the held-out validation parameters. For each parameter value and each time point we compute the RMS error of the full conserved state, and we average this quantity over the time horizon. Figure~\ref{fig-swe-fields} shows the predicted depth field alongside the solver reference and the pointwise error at representative times for a typical inner dam height. The network reproduces the outward-propagating circular bore and the inner rarefaction, and the largest discrepancies remain localized near the moving shock front where the solution is least smooth. Across the validation range the mean RMS error of the conserved state is $3.5 \times 10^{-3}$, and it generally increases with the inner dam height, from about $2.7 \times 10^{-3}$ near $h_{\mathrm{in}} = 1.7$ to about $6.0 \times 10^{-3}$ at $h_{\mathrm{in}} = 3.0$, as shown in Figure~\ref{fig-swe-error}. This growth reflects the stronger shock produced by larger initial imbalances, which is consistent with the localized errors observed at the moving front. These results confirm that the network learns a solver-consistent family of dam break solutions over the full range of inner dam heights without access to a labeled trajectory dataset during the solver-consistency phase.

\subsection{Three-dimensional Linear Elasticity}

The final stationary problem we study moves the solver-consistency
formulation into three dimensions and a vector-valued unknown. Linear
elasticity is a stringent test for two reasons. The solution is a
displacement field with three coupled components rather than a scalar,
and the reference solver is an unstructured finite-element method whose
operator the network must remain consistent with throughout training.
We adopt the clamped beam with three transverse holes that appears in
the NGSolve solid mechanics tutorial, a geometry whose stress
concentrations around the holes make the field nontrivial to reproduce.

The body occupies a box of dimensions $3.0 \times 0.6 \times 1.0$ with
three cylindrical holes of radius $0.25$ bored through its thickness at
even spacing along the long axis. The face at $x = 0$ is fully clamped
and the opposite face at $x = L_x$ carries a constant surface traction
$g = (0.3, 0, 0)$ aligned with the axis of the beam, with every other
surface left traction free. We use the linear strain
\begin{equation}
\varepsilon(u) = \tfrac{1}{2}(\nabla u + \nabla u^{\top}),
\end{equation}
and the Cauchy stress given by Hooke's law
\begin{equation}
\sigma(u) = 2\mu\,\varepsilon(u) + \lambda\,\mathrm{tr}\,\varepsilon(u)\,I.
\end{equation}
The displacement $u$ satisfies the weak equilibrium condition
\begin{equation}
  \int_{\Omega} \sigma(u) : \varepsilon(v)\,\mathrm{d}x
  = \int_{\Gamma_N} g \cdot v\,\mathrm{d}s
  \qquad \text{for all admissible } v ,
  \label{eq:elasticity-weak}
\end{equation}
subject to $u = 0$ on the clamped face. The Lam\'e parameters
$\mu = E / (2(1 + \nu))$ and
$\lambda = E\nu / ((1 + \nu)(1 - 2\nu))$ are fixed by Young's modulus
$E$ and Poisson's ratio $\nu$.

We hold $E = 1$ and treat Poisson's ratio as the conditioning parameter and sweep it over the range $[0.10, 0.45]$. The upper limit stays clear
of the incompressible limit $\nu \to 0.5$ where $\lambda$ diverges. The reference solver discretises
Eq.\ \eqref{eq:elasticity-weak} with second-order vector $H^1$ elements on a
curved tetrahedral mesh and solves the resulting system via the conjugate
gradients method with a BDDC preconditioner \citep{BrennerBDDC_2013}. Ground-truth displacements for evaluation come from running this solver to a tight tolerance of $10^{-8}$, while the
solver-consistency operator applied during training advances the
network's predicted field by only five preconditioned conjugate-gradient
iterations, which keeps the partial-correction behavior that the
\method objective depends on rather than collapsing it to a full solve.

\begin{figure}[h]
  \centering
  \begin{subfigure}[b]{0.4\linewidth}
    \centering
    \includegraphics[width=\linewidth]{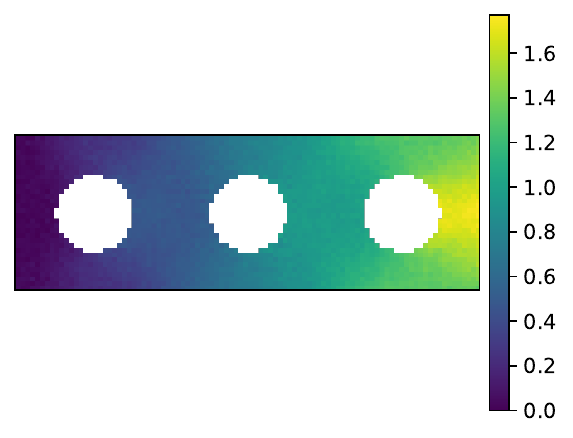}
    \caption{Predicted magnitude}
    \label{fig:elasticity-mag-pred}
  \end{subfigure}
  \hfill
  \begin{subfigure}[b]{0.4\linewidth}
    \centering
    \includegraphics[width=\linewidth]{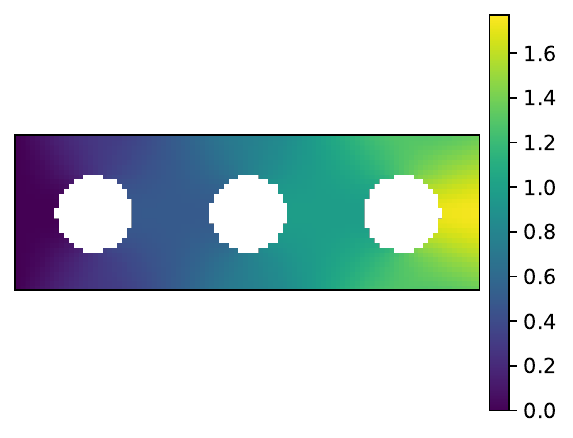}
    \caption{Reference magnitude}
    \label{fig:elasticity-mag-true}
  \end{subfigure}

  \begin{subfigure}[b]{0.4\linewidth}
    \centering
    \includegraphics[width=\linewidth]{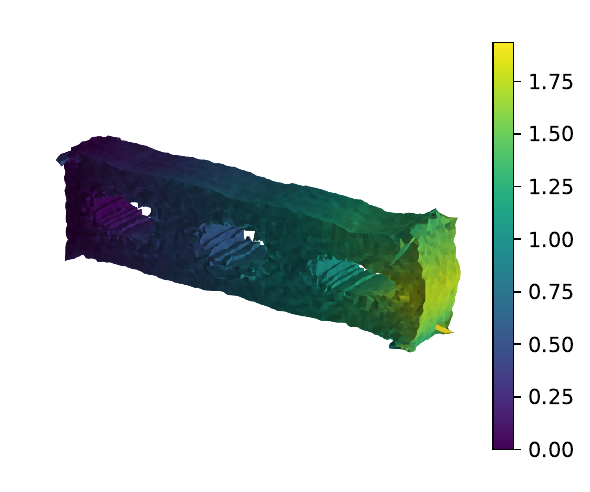}
    \caption{Predicted deformed body}
    \label{fig:elasticity-def-pred}
  \end{subfigure}
  \hfill
  \begin{subfigure}[b]{0.4\linewidth}
    \centering
    \includegraphics[width=\linewidth]{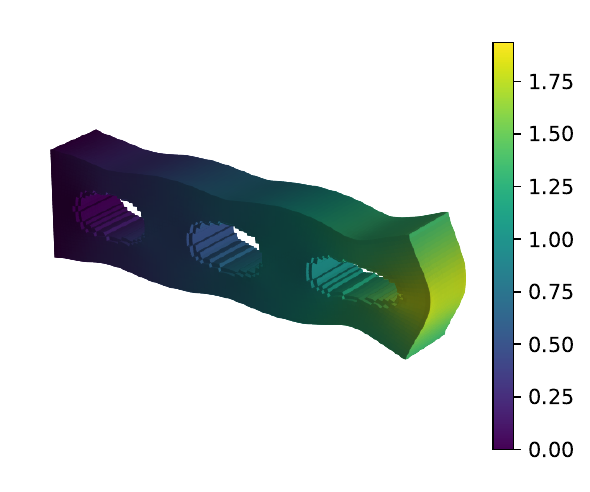}
    \caption{Reference deformed body}
    \label{fig:elasticity-def-true}
  \end{subfigure}
  \caption{Predicted and reference displacement for the clamped beam at
    Poisson's ratio $\nu = 0.30$. The top row shows the displacement
    magnitude on the central plane that passes through the three holes,
    and the bottom row shows the corresponding three-dimensional bodies
    warped by the predicted and reference displacements and coloured by
    magnitude, with the deformation scaled for visibility.
    The clamped face on the left carries zero displacement and the
    magnitude increases smoothly toward the loaded face on the right.
    The prediction matches the reference apart from a mild roughness on
    the deformed surface.}
  \label{fig:elasticity-fields}
\end{figure}

The network is a three-dimensional U-Net that reads two input channels,
a constant field carrying the normalized Poisson's ratio and a binary
mask marking the solid region, and predicts the three displacement
components on a $32 \times 16 \times 96$ grid that resolves the beam and
its holes, with roughly four fifths of the cells lying inside the body.
Training follows the steady-state \method procedure, advancing the
predicted field by the same five preconditioned conjugate-gradient steps
at each training step. We sample 256 values of Poisson's ratio at random
from $[0.10, 0.45]$ and train for 500 epochs at a batch size of 64. The
training loss is a mask-restricted mean-squared error between the
predicted field and its solver-advanced counterpart, evaluated only on
cells inside the body so that the holes and the surrounding void
contribute nothing.

Figure~\ref{fig:elasticity-fields} shows the predicted and reference
solutions at the representative value $\nu = 0.30$. On the central plane
that cuts through the three holes the predicted magnitude grows smoothly
from the clamped face toward the loaded face and reproduces the local
structure around each hole, and the three-dimensional bodies warped by
the two fields are nearly indistinguishable in shape, with the only
visible difference a mild high-frequency roughness on the predicted
surface. This instance lies close to the minimum of the error curve
discussed next.

\begin{figure}[h]
  \centering
  \includegraphics[width=0.5\linewidth]{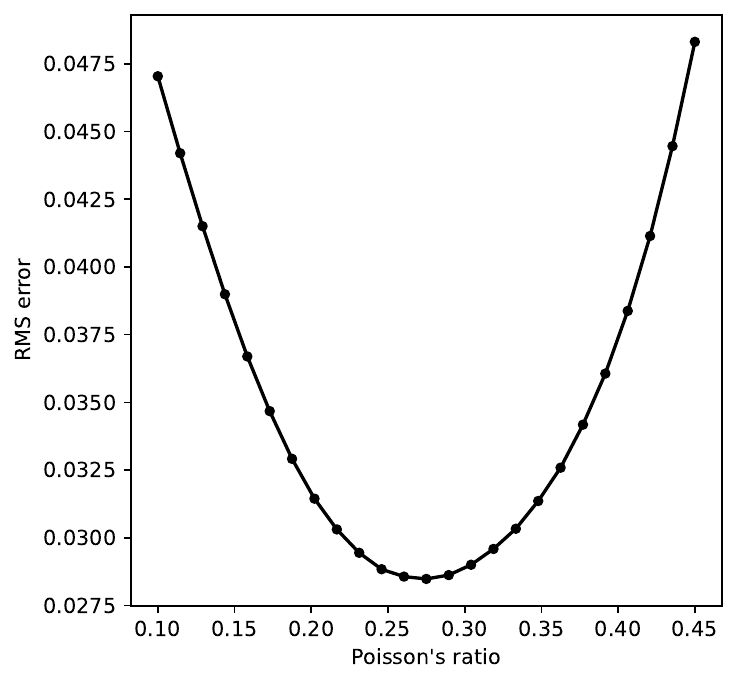}
  \caption{Root-mean-square error of the predicted displacement
    magnitude over the interior of the beam as a function of Poisson's
    ratio, evaluated at twenty-five values across the swept range. The
    error is smallest near the centre of the range and grows toward both
    ends, with the rise at the upper end reflecting the stiffening of the
    elastic response as the incompressible limit is approached.}
  \label{fig:elasticity-rms}
\end{figure}

To measure accuracy across the parameter range we evaluate the network
at twenty-five values of $\nu$ spanning $[0.10, 0.45]$ and report the
root-mean-square error of the displacement magnitude over the interior
cells, the same metric used throughout the preceding experiments.
Figure~\ref{fig:elasticity-rms} shows that the error stays small and
varies smoothly with $\nu$. It reaches a minimum of $2.85 \times 10^{-2}$
near the centre of the range at $\nu \approx 0.275$ and rises toward
both ends, reaching $4.70 \times 10^{-2}$ at $\nu = 0.10$ and
$4.83 \times 10^{-2}$ at $\nu = 0.45$, with the larger error at the
upper end consistent with the stiffening of the response as the
incompressible limit is approached. Averaged over the sweep the
root-mean-square error is $3.51 \times 10^{-2}$, which corresponds to
roughly four percent of the displacement magnitude. Obtaining this level
of agreement on a vector-valued field in three dimensions, with an
unstructured finite-element solver held in the training loop, shows that
the solver-consistency formulation carries over from the scalar and
two-dimensional problems to full three-dimensional solid mechanics
without modification.


\section{Conclusion}		\label{sec:conclusion}
We introduced \method, a solver-coupled learning framework for forward and inverse problems governed by partial differential equations. Rather than enforcing governing equations and boundary conditions through automatically differentiated residual penalties, \method uses a conventional numerical solver to generate state-dependent consistency targets from the neural network's own predictions. The solver can therefore be treated as a non-differentiable black-box operator, while the network learns an amortized parameter-to-solution map whose outputs are consistent with the solver's discrete dynamics, boundary conditions, and equilibrium structure.

We evaluated \method across finite-volume and finite-element discretizations spanning steady and transient systems, scalar and vector-valued fields, structured and unstructured domains, and both smooth and discontinuous solution regimes. The steady-state experiments included the two-dimensional Fokker--Planck equation, the high-Reynolds-number lid-driven cavity, and three-dimensional linear elasticity on a perforated geometry. The transient experiments included variable-coefficient advection, radial shallow-water dam-break dynamics, and incompressible flow past a cylinder with periodic vortex shedding. Across these problems, the learned models reproduced the corresponding numerical reference solutions over their prescribed parameter ranges. The inverse experiments further showed that unknown transport and flow parameters could be recovered from noisy observations by differentiating through the learned surrogate while retaining solver-consistency in the forward model.

These results demonstrate that solver-consistency training can accommodate multiple numerical discretizations, complex geometries, nonlinear transport, shock-like features, oscillatory flow, and three-dimensional mechanics without differentiating through the solver or constructing problem-specific PDE residual losses. At the same time, \method inherits the accuracy, stability requirements, invariant structure, and computational cost of the underlying numerical method. Training may therefore require solver-admissible initialization, normalization, or warm-starting, particularly for strongly nonlinear, nearly incompressible, or discontinuous systems. Within these limitations, the present experiments indicate that black-box solver-coupled learning is a flexible approach for constructing reusable scientific surrogates and solving parameter-identification problems across computational physics and engineering.

%
%

\bibliography{satish}

@Book{panton,
  author    = {Panton, Ronald L.},
  title     = {Incompressible Flow},
  publisher = {John Wiley \& Sons, Ltd},
  year      = {2013},
  isbn      = {9781118013434},
}

@Book{LarsenFEM,
author={Mats G. Larson, Fredrik Bengzon},
title={The Finite Element Method: Theory, Implementation, and Applications},
year={2013},
publisher={Springer Berlin, Heidelberg},
}

@book{LeVequeFVM, place={Cambridge}, series={Cambridge Texts in Applied Mathematics}, title={Finite Volume Methods for Hyperbolic Problems}, publisher={Cambridge University Press}, author={LeVeque, Randall J.}, year={2002}, collection={Cambridge Texts in Applied Mathematics}}

@Book{BrennerScottFEM,
author={Susanne C. Brenner, L. Ridgway Scott},
title={The Mathematical Theory of Finite Element Methods},
year={2008},
publisher={Springer New York, NY},
}

@book{LevequeFDM,
author = {LeVeque, Randall J.},
title = {Finite Difference Methods for Ordinary and Partial Differential Equations},
publisher = {Society for Industrial and Applied Mathematics},
year = {2007},
doi = {10.1137/1.9780898717839},
address = {},
edition   = {},
URL = {https://epubs.siam.org/doi/abs/10.1137/1.9780898717839},
eprint = {https://epubs.siam.org/doi/pdf/10.1137/1.9780898717839}
}

@Book{LarssonThomeeNumPDE,
author={Stig Larsson, Vidar Thomée},
title={Partial Differential Equations with Numerical Methods},
year={2003},
publisher={Springer Berlin, Heidelberg},
}

@book{JacksonEM,
author = {Jackson, J. D.},
publisher = {John Wiley \& Sons, Ltd},
isbn = {9783527600434},
title = {Electrodynamics, Classical},
booktitle = {Encyclopedia of Applied Physics},
doi = {https://doi.org/10.1002/3527600434.eap109},
url = {https://onlinelibrary.wiley.com/doi/abs/10.1002/3527600434.eap109},
eprint = {https://onlinelibrary.wiley.com/doi/pdf/10.1002/3527600434.eap109},
year = {2003}
}

@book{dym2013solid,
  title        = {Solid Mechanics: A Variational Approach, Augmented Edition},
  author       = {Clive L. Dym and Irving H. Shames},
  year         = {2013},
  publisher    = {Springer New York},
  address      = {New York, NY},
  isbn         = {978-1-4614-6033-6},
  doi          = {10.1007/978-1-4614-6034-3},
  url          = {https://link.springer.com/book/10.1007/978-1-4614-6034-3}
}

@book{evans10,
  address = {Providence, R.I.},
  author = {Evans, Lawrence C.},
  isbn = {9780821849743 0821849743},
  publisher = {American Mathematical Society},
  refid = {465190110},
  title = {Partial differential equations},
  year = 2010
}

@article{RAISSI2019,
title = {Physics-informed neural networks: A deep learning framework for solving forward and inverse problems involving nonlinear partial differential equations},
journal = {Journal of Computational Physics},
volume = {378},
pages = {686-707},
year = {2019},
issn = {0021-9991},
doi = {https://doi.org/10.1016/j.jcp.2018.10.045},
url = {https://www.sciencedirect.com/science/article/pii/S0021999118307125},
author = {M. Raissi and P. Perdikaris and G.E. Karniadakis}
}

@article{
RaissiHiddenFluid2020,
author = {Maziar Raissi  and Alireza Yazdani  and George Em Karniadakis },
title = {Hidden fluid mechanics: Learning velocity and pressure fields from flow visualizations},
journal = {Science},
volume = {367},
number = {6481},
pages = {1026-1030},
year = {2020},
doi = {10.1126/science.aaw4741},
URL = {https://www.science.org/doi/abs/10.1126/science.aaw4741},
eprint = {https://www.science.org/doi/pdf/10.1126/science.aaw4741}}

@article{ZhaoZhangLouWang2024,
    author = {Zhao, Chi and Zhang, Feifei and Lou, Wenqiang and Wang, Xi and Yang, Jianyong},
    title = {A comprehensive review of advances in physics-informed neural networks and their applications in complex fluid dynamics},
    journal = {Physics of Fluids},
    volume = {36},
    number = {10},
    pages = {101301},
    year = {2024},
    month = {10},
    issn = {1070-6631},
    doi = {10.1063/5.0226562},
    url = {https://doi.org/10.1063/5.0226562},
    eprint = {https://pubs.aip.org/aip/pof/article-pdf/doi/10.1063/5.0226562/20203311/101301\_1\_5.0226562.pdf},
}

@article{CaoSongZhang2024,
    author = {Cao, Wenbo and Song, Jiahaoand Zhang, Weiwei},
    title = {A solver for subsonic flow around airfoils based on physics-informed neural networks and mesh transformation},
    journal = {Physics of Fluids},
    volume = {36},
    number = {2},
    pages = {027134},
    year = {2024},
    month = {02},
    issn = {1070-6631},
    doi = {10.1063/5.0188665},
    url = {https://doi.org/10.1063/5.0188665},
    eprint = {https://pubs.aip.org/aip/pof/article-pdf/doi/10.1063/5.0188665/19692344/027134\_1\_5.0188665.pdf},
}

@article{CaoZhang2025,
title = {An analysis and solution of ill-conditioning in physics-informed neural networks},
journal = {Journal of Computational Physics},
volume = {520},
pages = {113494},
year = {2025},
issn = {0021-9991},
doi = {https://doi.org/10.1016/j.jcp.2024.113494},
url = {https://www.sciencedirect.com/science/article/pii/S0021999124007423},
author = {Wenbo Cao and Weiwei Zhang}
}

@article{Ertuck2005,
author = {Erturk, E. and Corke, T. C. and Gökçöl, C.},
title = {Numerical solutions of 2-D steady incompressible driven cavity flow at high Reynolds numbers},
journal = {International Journal for Numerical Methods in Fluids},
volume = {48},
number = {7},
pages = {747-774},
doi = {https://doi.org/10.1002/fld.953},
url = {https://onlinelibrary.wiley.com/doi/abs/10.1002/fld.953},
eprint = {https://onlinelibrary.wiley.com/doi/pdf/10.1002/fld.953},
year = {2005}
}

@article{ARUN2015,
title = {Analysis of flow behaviour in a two sided lid driven cavity using lattice boltzmann technique},
journal = {Alexandria Engineering Journal},
volume = {54},
number = {4},
pages = {795-806},
year = {2015},
issn = {1110-0168},
doi = {https://doi.org/10.1016/j.aej.2015.06.005},
url = {https://www.sciencedirect.com/science/article/pii/S1110016815000976},
author = {S. Arun and A. Satheesh}
}

@article{ARUMUGAPERUMAL2011,
title = {Multiplicity of steady solutions in two-dimensional lid-driven cavity flows by Lattice Boltzmann Method},
journal = {Computers \& Mathematics with Applications},
volume = {61},
number = {12},
pages = {3711-3721},
year = {2011},
note = {Mesoscopic Methods for Engineering and Science — Proceedings of ICMMES-09},
issn = {0898-1221},
doi = {https://doi.org/10.1016/j.camwa.2010.03.053},
url = {https://www.sciencedirect.com/science/article/pii/S0898122110002427},
author = {D. {Arumuga Perumal} and Anoop K. Dass}
}

@misc{ngsolve2025,
  author       = {Schöberl, Joachim and others},
  title        = {NGSolve: A finite element library},
  year         = {2025},
  howpublished = {\url{https://github.com/NGSolve/ngsolve}},
  note         = {Accessed: 2025-05-05}
}

@InProceedings{Ronneberger2015,
author="Ronneberger, Olaf
and Fischer, Philipp
and Brox, Thomas",
editor="Navab, Nassir
and Hornegger, Joachim
and Wells, William M.
and Frangi, Alejandro F.",
title="U-Net: Convolutional Networks for Biomedical Image Segmentation",
booktitle="Medical Image Computing and Computer-Assisted Intervention -- MICCAI 2015",
year="2015",
publisher="Springer International Publishing",
address="Cham",
pages="234--241",
isbn="978-3-319-24574-4"
}

@book{Boscarino2024,
author = {Boscarino, Sebastiano and Pareschi, Lorenzo and Russo, Giovanni},
title = {Implicit-Explicit Methods for Evolutionary Partial Differential Equations},
publisher = {Society for Industrial and Applied Mathematics},
year = {2024},
doi = {10.1137/1.9781611978209},
address = {Philadelphia, PA},
edition   = {},
URL = {https://epubs.siam.org/doi/abs/10.1137/1.9781611978209},
eprint = {https://epubs.siam.org/doi/pdf/10.1137/1.9781611978209}
}

@article{Zheng,
author = {Zheng, Bohong},
year = {2023},
month = {12},
pages = {645-651},
title = {Ordinary Differential Equation and Its Application},
volume = {72},
journal = {Highlights in Science, Engineering and Technology},
doi = {10.54097/rnnev212}
}

@article{YANG2025103997,
title = {PINN neural network method for solving the forward and inverse problem of time-fractional telegraph equation},
journal = {Results in Engineering},
volume = {25},
pages = {103997},
year = {2025},
issn = {2590-1230},
doi = {https://doi.org/10.1016/j.rineng.2025.103997},
url = {https://www.sciencedirect.com/science/article/pii/S2590123025000854},
author = {Fan Yang and Hao Liu and Xiao-Xiao Li and Jian-Xiong Cao}
}

@article{GAO2022114502,
title = {Physics-informed graph neural Galerkin networks: A unified framework for solving PDE-governed forward and inverse problems},
journal = {Computer Methods in Applied Mechanics and Engineering},
volume = {390},
pages = {114502},
year = {2022},
issn = {0045-7825},
doi = {https://doi.org/10.1016/j.cma.2021.114502},
url = {https://www.sciencedirect.com/science/article/pii/S0045782521007076},
author = {Han Gao and Matthew J. Zahr and Jian-Xun Wang}
}

@inproceedings{10.5555/3692070.3693785,
author = {Rathore, Pratik and Lei, Weimu and Frangella, Zachary and Lu, Lu and Udell, Madeleine},
title = {Challenges in training PINNs: a loss landscape perspective},
year = {2024},
publisher = {JMLR.org},
booktitle = {Proceedings of the 41st International Conference on Machine Learning},
articleno = {1715},
numpages = {33},
location = {Vienna, Austria},
series = {ICML'24}
}

@article{SONG2024112781,
title = {Loss-attentional physics-informed neural networks},
journal = {Journal of Computational Physics},
volume = {501},
pages = {112781},
year = {2024},
issn = {0021-9991},
doi = {https://doi.org/10.1016/j.jcp.2024.112781},
url = {https://www.sciencedirect.com/science/article/pii/S0021999124000305},
author = {Yanjie Song and He Wang and He Yang and Maria Luisa Taccari and Xiaohui Chen}
}

@article{LEE2024117000,
title = {Anti-derivatives approximator for enhancing physics-informed neural networks},
journal = {Computer Methods in Applied Mechanics and Engineering},
volume = {426},
pages = {117000},
year = {2024},
issn = {0045-7825},
doi = {https://doi.org/10.1016/j.cma.2024.117000},
url = {https://www.sciencedirect.com/science/article/pii/S0045782524002561},
author = {Jeongsu Lee}
}

@article{JAGTAP2020113028,
title = {Conservative physics-informed neural networks on discrete domains for conservation laws: Applications to forward and inverse problems},
journal = {Computer Methods in Applied Mechanics and Engineering},
volume = {365},
pages = {113028},
year = {2020},
issn = {0045-7825},
doi = {https://doi.org/10.1016/j.cma.2020.113028},
url = {https://www.sciencedirect.com/science/article/pii/S0045782520302127},
author = {Ameya D. Jagtap and Ehsan Kharazmi and George Em Karniadakis}
}

@article{WANG2024116813,
title = {Respecting causality for training physics-informed neural networks},
journal = {Computer Methods in Applied Mechanics and Engineering},
volume = {421},
pages = {116813},
year = {2024},
issn = {0045-7825},
doi = {https://doi.org/10.1016/j.cma.2024.116813},
url = {https://www.sciencedirect.com/science/article/pii/S0045782524000690},
author = {Sifan Wang and Shyam Sankaran and Paris Perdikaris}
}

@article{wangetal,
author = {Wang, Sifan and Teng, Yujun and Perdikaris, Paris},
year = {2021},
month = {09},
pages = {A3055-A3081},
title = {Understanding and Mitigating Gradient Flow Pathologies in Physics-Informed Neural Networks},
volume = {43},
journal = {SIAM Journal on Scientific Computing},
doi = {10.1137/20M1318043}
}

@article{JIN2024107887,
title = {Fourier warm start for physics-informed neural networks},
journal = {Engineering Applications of Artificial Intelligence},
volume = {132},
pages = {107887},
year = {2024},
issn = {0952-1976},
doi = {https://doi.org/10.1016/j.engappai.2024.107887},
url = {https://www.sciencedirect.com/science/article/pii/S0952197624000459},
author = {Ge Jin and Jian Cheng Wong and Abhishek Gupta and Shipeng Li and Yew-Soon Ong}
}

@Article{app14083204,
AUTHOR = {Mustajab, Abdul Hannan and Lyu, Hao and Rizvi, Zarghaam and Wuttke, Frank},
TITLE = {Physics-Informed Neural Networks for High-Frequency and Multi-Scale Problems Using Transfer Learning},
JOURNAL = {Applied Sciences},
VOLUME = {14},
YEAR = {2024},
NUMBER = {8},
ARTICLE-NUMBER = {3204},
URL = {https://www.mdpi.com/2076-3417/14/8/3204},
ISSN = {2076-3417},
DOI = {10.3390/app14083204}
}

@article{YE2024111006,
title = {Modeling of the hysteretic behavior of nonlinear particle damping by Fourier neural network with transfer learning},
journal = {Mechanical Systems and Signal Processing},
volume = {208},
pages = {111006},
year = {2024},
issn = {0888-3270},
doi = {https://doi.org/10.1016/j.ymssp.2023.111006},
url = {https://www.sciencedirect.com/science/article/pii/S0888327023009147},
author = {Xin Ye and Yi-Qing Ni and Wai Kei Ao and Lei Yuan}
}

@article{Zhang2024,
  author       = {Zhang, Enrui and Kahana, Adar and Kopaničáková, Alena and Turkel, Eli and Ranade, Rishikesh and Pathak, Jay and Karniadakis, George Em},
  title        = {Blending neural operators and relaxation methods in PDE numerical solvers},
  journal      = {Nature Machine Intelligence},
  volume       = {6},
  number       = {11},
  pages        = {1303--1313},
  year         = {2024},
  month        = nov,
  doi          = {10.1038/s42256-024-00910-x},
  url          = {https://doi.org/10.1038/s42256-024-00910-x},
  issn         = {2522-5839}
}

@article{Oommen2024,
  author       = {Oommen, Vivek and Shukla, Khemraj and Desai, Saaketh and Dingreville, Rémi and Karniadakis, George Em},
  title        = {Rethinking materials simulations: Blending direct numerical simulations with neural operators},
  journal      = {npj Computational Materials},
  volume       = {10},
  number       = {1},
  pages        = {145},
  year         = {2024},
  month        = jul,
  doi          = {10.1038/s41524-024-01319-1},
  url          = {https://doi.org/10.1038/s41524-024-01319-1},
  issn         = {2057-3960}
}

@article{MERLET2024106316,
title = {Mixing neural networks, continuation and symbolic computation to solve parametric systems of non linear equations},
journal = {Neural Networks},
volume = {176},
pages = {106316},
year = {2024},
issn = {0893-6080},
doi = {https://doi.org/10.1016/j.neunet.2024.106316},
url = {https://www.sciencedirect.com/science/article/pii/S0893608024002405},
author = {J.-P. Merlet}
}

@inproceedings{NEURIPS2018_69386f6b,
 author = {Chen, Ricky T. Q. and Rubanova, Yulia and Bettencourt, Jesse and Duvenaud, David K},
 booktitle = {Advances in Neural Information Processing Systems},
 editor = {S. Bengio and H. Wallach and H. Larochelle and K. Grauman and N. Cesa-Bianchi and R. Garnett},
 pages = {},
 publisher = {Curran Associates, Inc.},
 title = {Neural Ordinary Differential Equations},
 url = {https://proceedings.neurips.cc/paper_files/paper/2018/file/69386f6bb1dfed68692a24c8686939b9-Paper.pdf},
 volume = {31},
 year = {2018}
}

@article{Rackauckas,
author = {Rackauckas, Chris and Ma, Yingbo and Martensen, Julius and Warner, Collin and Zubov, Kirill and Supekar, Rohit and Skinner, Dominic and Ramadhan, Ali},
year = {2020},
month = {01},
pages = {},
journal={ArXiv},
title = {Universal Differential Equations for Scientific Machine Learning},
doi = {10.21203/rs.3.rs-55125/v1}
}

@article{Egan2024,
  author       = {Egan, Kevin and Li, Weizhen and Carvalho, Rui},
  title        = {Automatically discovering ordinary differential equations from data with sparse regression},
  journal      = {Communications Physics},
  volume       = {7},
  number       = {1},
  pages        = {20},
  year         = {2024},
  month        = jan,
  doi          = {10.1038/s42005-023-01516-2},
  url          = {https://doi.org/10.1038/s42005-023-01516-2},
  issn         = {2399-3650}
}

@article{PU2023107051,
title = {Data-driven forward-inverse problems for Yajima–Oikawa system using deep learning with parameter regularization},
journal = {Communications in Nonlinear Science and Numerical Simulation},
volume = {118},
pages = {107051},
year = {2023},
issn = {1007-5704},
doi = {https://doi.org/10.1016/j.cnsns.2022.107051},
url = {https://www.sciencedirect.com/science/article/pii/S100757042200538X},
author = {Jun-Cai Pu and Yong Chen}
}

@article{MEIDANI2021113831,
title = {Data-driven identification of 2D Partial Differential Equations using extracted physical features},
journal = {Computer Methods in Applied Mechanics and Engineering},
volume = {381},
pages = {113831},
year = {2021},
issn = {0045-7825},
doi = {https://doi.org/10.1016/j.cma.2021.113831},
url = {https://www.sciencedirect.com/science/article/pii/S0045782521001687},
author = {Kazem Meidani and Amir {Barati Farimani}}
}

@article{Xu2023,
  author       = {Xu, H. and Zeng, J. and Zhang, D.},
  title        = {Discovery of Partial Differential Equations from Highly Noisy and Sparse Data with Physics-Informed Information Criterion},
  journal      = {Research},
  year         = {2023},
  volume       = {6},
  pages        = {0147},
  month        = may,
  doi          = {10.34133/research.0147},
  pmid         = {37214196},
  pmcid        = {PMC10198462}
}

@article{BRUNTON2016710,
title = {Sparse Identification of Nonlinear Dynamics with Control (SINDYc)},
journal = {IFAC-PapersOnLine},
volume = {49},
number = {18},
pages = {710-715},
year = {2016},
note = {10th IFAC Symposium on Nonlinear Control Systems NOLCOS 2016},
issn = {2405-8963},
doi = {https://doi.org/10.1016/j.ifacol.2016.10.249},
url = {https://www.sciencedirect.com/science/article/pii/S2405896316318298},
author = {Steven L. Brunton and Joshua L. Proctor and J. Nathan Kutz}
}

@article{MaziarRaissi,
author = {Raissi, Maziar},
year = {2018},
month = {01},
pages = {},
title = {Deep Hidden Physics Models: Deep Learning of Nonlinear Partial Differential Equations},
volume = {19},
journal = {Journal of Machine Learning Research},
doi = {10.48550/arXiv.1801.06637}
}

@article{LONG2019108925,
title = {PDE-Net 2.0: Learning PDEs from data with a numeric-symbolic hybrid deep network},
journal = {Journal of Computational Physics},
volume = {399},
pages = {108925},
year = {2019},
issn = {0021-9991},
doi = {https://doi.org/10.1016/j.jcp.2019.108925},
url = {https://www.sciencedirect.com/science/article/pii/S0021999119306308},
author = {Zichao Long and Yiping Lu and Bin Dong}
}

@book{Blazek2015,
  author    = {Blazek, J.},
  title     = {Computational Fluid Dynamics: Principles and Applications},
  publisher = {Elsevier},
  year      = {2015},
  edition   = {3},
  isbn      = {9780127999144}
}

@book{LeVeque2007,
  author    = {Randall J. LeVeque},
  title     = {Finite Difference Methods for Ordinary and Partial Differential Equations: Steady-State and Time-Dependent Problems},
  publisher = {Society for Industrial and Applied Mathematics},
  year      = {2007},
  address   = {Philadelphia, PA},
  isbn      = {978-0-898716-29-0},
  doi       = {10.1137/1.9780898717839}
}

@article{Boussaid,
author = {Boussaid, Samira and Hilhorst, Danielle and Nguyen, Thanh},
year = {2015},
month = {03},
pages = {39-59},
title = {Convergence to steady state for the solutions of a nonlocal reaction-diffusion equation},
volume = {4},
journal = {Evolution Equations and Control Theory},
doi = {10.3934/eect.201.5.4.39}
}

@inproceedings{
marwah2023deep,
title={Deep Equilibrium Based Neural Operators for Steady-State {PDE}s},
author={Tanya Marwah and Ashwini Pokle and J Zico Kolter and Zachary Chase Lipton and Jianfeng Lu and Andrej Risteski},
booktitle={Thirty-seventh Conference on Neural Information Processing Systems},
year={2023},
url={https://openreview.net/forum?id=v6YzxwJlQn}
}

@book{Granas2003,
  author    = {Andrzej Granas and James Dugundji},
  title     = {Fixed Point Theory},
  publisher = {Springer},
  year      = {2003},
  series    = {Springer Monographs in Mathematics},
  isbn      = {978-0387001735},
  url       = {https://books.google.com/books/about/Fixed_Point_Theory.html?id=4_iJAoLSq3cC}
}

@article{Ji2021,
author = {Ji, Weiqi and Qiu, Weilun and Shi, Zhiyu and Pan, Shaowu and Deng, Sili},
title = {Stiff-PINN: Physics-Informed Neural Network for Stiff Chemical Kinetics},
journal = {The Journal of Physical Chemistry A},
volume = {125},
number = {36},
pages = {8098-8106},
year = {2021},
doi = {10.1021/acs.jpca.1c05102},
    note ={PMID: 34463510},

URL = { 
    
        https://doi.org/10.1021/acs.jpca.1c05102
    
    

},
eprint = { 
    
        https://doi.org/10.1021/acs.jpca.1c05102
    
    

}

}

@Article{Huang2022,
author={Huang, Yunfei
and Mabrouk, Youssef
and Gompper, Gerhard
and Sabass, Benedikt},
title={Sparse inference and active learning of stochastic differential equations from data},
journal={Scientific Reports},
year={2022},
month={Dec},
day={15},
volume={12},
number={1},
pages={21691},
issn={2045-2322},
doi={10.1038/s41598-022-25638-9},
url={https://doi.org/10.1038/s41598-022-25638-9}
}

@book{Tsai2018,
	author = {Tsai, Tai-Peng},
	title = {Lectures on Navier-Stokes equations },
	publisher = {American Mathematical Society,},
	year = {2018.},
	series = {Graduate Studies in Mathematics, },
	address = {Providence, R.I. :},
	url = {https://doi.org/10.1090/gsm/192}
}

@Book{Bathe2006,
    author      = {Bathe, Klaus-J{\"u}rgen},
    title       = {Finite Element Procedures},
    publisher   = {Klaus-Jurgen Bathe},
    year        = {2006}
}

@article{CHEN2023,
title = {Deep-OSG: Deep learning of operators in semigroup},
journal = {Journal of Computational Physics},
volume = {493},
pages = {112498},
year = {2023},
issn = {0021-9991},
doi = {https://doi.org/10.1016/j.jcp.2023.112498},
url = {https://www.sciencedirect.com/science/article/pii/S0021999123005934},
author = {Junfeng Chen and Kailiang Wu}
}

@book{gustafsson2013time,
  title     = {Time-Dependent Problems and Difference Methods},
  author    = {Gustafsson, Bertil and Kreiss, Heinz-Otto and Oliger, Joseph},
  year      = {2013},
  edition   = {2},
  publisher = {John Wiley \& Sons},
  address   = {Hoboken, NJ},
  isbn      = {978-0-470-90056-7},
  series    = {Pure and Applied Mathematics: A Wiley Series of Texts, Monographs and Tracts}
}

@article{GHIA1982,
title = {High-Re solutions for incompressible flow using the Navier-Stokes equations and a multigrid method},
journal = {Journal of Computational Physics},
volume = {48},
number = {3},
pages = {387-411},
year = {1982},
issn = {0021-9991},
doi = {https://doi.org/10.1016/0021-9991(82)90058-4},
url = {https://www.sciencedirect.com/science/article/pii/0021999182900584},
author = {U Ghia and K.N Ghia and C.T Shin}
}

@misc{AdamWpaper,
      title={Decoupled Weight Decay Regularization}, 
      author={Ilya Loshchilov and Frank Hutter},
      year={2019},
      eprint={1711.05101},
      archivePrefix={arXiv},
      primaryClass={cs.LG},
      url={https://arxiv.org/abs/1711.05101}, 
}

@article{XLB,
title = {XLB: A differentiable massively parallel lattice Boltzmann library in Python},
journal = {Computer Physics Communications},
volume = {300},
pages = {109187},
year = {2024},
issn = {0010-4655},
doi = {https://doi.org/10.1016/j.cpc.2024.109187},
url = {https://www.sciencedirect.com/science/article/pii/S0010465524001103},
author = {Mohammadmehdi Ataei and Hesam Salehipour}
}

@article{Ginzburg2008,
author = {Ginzburg, Irina},
year = {2008},
month = {01},
pages = {427-478},
title = {Two-relaxation-time Lattice Boltzmann scheme: about parametrization, velocity, pressure and mixed boundary conditions},
volume = {3},
journal = {Communications in Computational Physics}
}

@article{HOSSEINI2023,
title = {Entropic lattice Boltzmann methods: A review},
journal = {Computers \& Fluids},
volume = {259},
pages = {105884},
year = {2023},
issn = {0045-7930},
doi = {https://doi.org/10.1016/j.compfluid.2023.105884},
url = {https://www.sciencedirect.com/science/article/pii/S0045793023001093},
author = {S.A. Hosseini and M. Atif and S. Ansumali and I.V. Karlin}
}

@incollection{DEVAUCORBEIL2020,
title = {Chapter Two - Material point method after 25 years: Theory, implementation, and applications},
booktitle = {Chapter Two - Material point method after 25 years: Theory, implementation, and applications},
editor = {Stéphane P.A. Bordas and Daniel S. Balint},
series = {Advances in Applied Mechanics},
publisher = {Elsevier},
volume = {53},
pages = {185-398},
year = {2020},
issn = {0065-2156},
doi = {https://doi.org/10.1016/bs.aams.2019.11.001},
url = {https://www.sciencedirect.com/science/article/pii/S0065215619300146},
author = {Alban {de Vaucorbeil} and Vinh Phu Nguyen and Sina Sinaie and Jian Ying Wu}
}

@book{nguyen2023material,
  title        = {The Material Point Method},
  subtitle     = {Theory, Implementations and Applications},
  author       = {Vinh Phu Nguyen and Alban de Vaucoubeil and Stephane Bordas},
  year         = {2023},
  publisher    = {Springer Cham},
  series       = {Scientific Computation},
  isbn         = {978-3-031-24069-0},
  doi          = {10.1007/978-3-031-24070-6},
  url          = {https://doi.org/10.1007/978-3-031-24070-6},
  edition      = {1},
  pages        = {XV, 467},
  eisbn        = {978-3-031-24070-6},
  issn         = {1434-8322},
  eissn        = {2198-2589},
  note         = {Hardcover published: 12 April 2023; eBook published: 11 April 2023; Softcover published: 13 April 2024}
}

@inproceedings{he2015delving,
  title={Delving deep into rectifiers: Surpassing human-level performance on ImageNet classification},
  author={He, Kaiming and Zhang, Xiangyu and Ren, Shaoqing and Sun, Jian},
  booktitle={Proceedings of the IEEE International Conference on Computer Vision (ICCV)},
  pages={1026--1034},
  year={2015},
  doi={10.1109/ICCV.2015.123}
}

@Article{Lu2021,
author={Lu, Lu
and Jin, Pengzhan
and Pang, Guofei
and Zhang, Zhongqiang
and Karniadakis, George Em},
title={Learning nonlinear operators via DeepONet based on the universal approximation theorem of operators},
journal={Nature Machine Intelligence},
year={2021},
month={Mar},
day={01},
volume={3},
number={3},
pages={218-229},
issn={2522-5839},
doi={10.1038/s42256-021-00302-5},
url={https://doi.org/10.1038/s42256-021-00302-5}
}

@article{fno2020,
  author       = {Zongyi Li and
                  Nikola B. Kovachki and
                  Kamyar Azizzadenesheli and
                  Burigede Liu and
                  Kaushik Bhattacharya and
                  Andrew M. Stuart and
                  Anima Anandkumar},
  title        = {Fourier Neural Operator for Parametric Partial Differential Equations},
  journal      = {CoRR},
  volume       = {abs/2010.08895},
  year         = {2020},
  url          = {https://arxiv.org/abs/2010.08895},
  eprinttype    = {arXiv},
  eprint       = {2010.08895},
  bibsource    = {dblp computer science bibliography, https://dblp.org}
}

@misc{bai2019deepequilibriummodels,
      title={Deep Equilibrium Models}, 
      author={Shaojie Bai and J. Zico Kolter and Vladlen Koltun},
      year={2019},
      eprint={1909.01377},
      archivePrefix={arXiv},
      primaryClass={cs.LG},
      url={https://arxiv.org/abs/1909.01377}, 
}

@misc{bai2020multiscaledeepequilibriummodels,
      title={Multiscale Deep Equilibrium Models}, 
      author={Shaojie Bai and Vladlen Koltun and J. Zico Kolter},
      year={2020},
      eprint={2006.08656},
      archivePrefix={arXiv},
      primaryClass={cs.LG},
      url={https://arxiv.org/abs/2006.08656}, 
}

@misc{winston2021monotoneoperatorequilibriumnetworks,
      title={Monotone operator equilibrium networks}, 
      author={Ezra Winston and J. Zico Kolter},
      year={2021},
      eprint={2006.08591},
      archivePrefix={arXiv},
      primaryClass={cs.LG},
      url={https://arxiv.org/abs/2006.08591}, 
}

@article{FiPy:2009, author = {Jonathan E. Guyer and Daniel Wheeler and James A. Warren}, title = {{FiPy}: Partial Differential Equations with {P}ython}, publisher = {IEEE}, year = 2009, journal = {Computing in Science \& Engineering}, volume = 11, number = 3, pages = {6-15}, url = {http://www.ctcms.nist.gov/fipy}, doi = {10.1109/MCSE.2009.52}, }

@misc{loshchilov2017sgdrstochasticgradientdescent,
      title={SGDR: Stochastic Gradient Descent with Warm Restarts}, 
      author={Ilya Loshchilov and Frank Hutter},
      year={2017},
      eprint={1608.03983},
      archivePrefix={arXiv},
      primaryClass={cs.LG},
      url={https://arxiv.org/abs/1608.03983}, 
}

@mastersthesis{haughey2009boundless,
  title        = {Boundless Fluids Using the Lattice-Boltzmann Method},
  author       = {Haughey, Kyle J.},
  school       = {California Polytechnic State University, San Luis Obispo},
  year         = {2009},
  month        = jun,
  type         = {Master’s Thesis},
  address      = {San Luis Obispo, CA},
  doi          = {10.15368/theses.2009.73},
  url          = {https://digitalcommons.calpoly.edu/theses/117/},
  department   = {Computer Science},
  advisor      = {Zoe J. Wood}
}

@misc{taghikhani2025neuralinitializednewtonacceleratingnonlinear,
      title={Neural-Initialized Newton: Accelerating Nonlinear Finite Elements via Operator Learning}, 
      author={Kianoosh Taghikhani and Yusuke Yamazaki and Jerry Paul Varghese and Markus Apel and Reza Najian Asl and Shahed Rezaei},
      year={2025},
      eprint={2511.06802},
      archivePrefix={arXiv},
      primaryClass={cs.LG},
      url={https://arxiv.org/abs/2511.06802}, 
}

@misc{wang2026pretrainfiniteelementmethod,
      title={Pretrain Finite Element Method: A Pretraining and Warm-start Framework for PDEs via Physics-Informed Neural Operators}, 
      author={Yizheng Wang and Zhongkai Hao and Mohammad Sadegh Eshaghi and Cosmin Anitescu and Xiaoying Zhuang and Timon Rabczuk and Yinghua Liu},
      year={2026},
      eprint={2601.03086},
      archivePrefix={arXiv},
      primaryClass={math.NA},
      url={https://arxiv.org/abs/2601.03086}, 
}

@misc{clawpack,
    title={Clawpack software},
    author={{Clawpack Development Team}},
    url={http://www.clawpack.org},
    note={Version 5.14.0},
    doi={https://doi.org/10.5281/zenodo.18382457},
    year={2026}}

@article{Albarede_Provansal_1995, title={Quasi-periodic cylinder wakes and the Ginzburg–Landau model}, volume={291}, DOI={10.1017/S0022112095002679}, journal={Journal of Fluid Mechanics}, author={Albarède, Pierre and Provansal, Michel}, year={1995}, pages={191–222}}

@article{Monkewitz1996,
    author = {Monkewitz, P. A. and Williamson, C. H. K. and Miller, G. D.},
    title = {Phase dynamics of Kármán vortices in cylinder wakes},
    journal = {Physics of Fluids},
    volume = {8},
    number = {1},
    pages = {91-96},
    year = {1996},
    month = {01},
    issn = {1070-6631},
    doi = {10.1063/1.868817},
    url = {https://doi.org/10.1063/1.868817},
    eprint = {https://pubs.aip.org/aip/pof/article-pdf/8/1/91/19248032/91_1_online.pdf},
}

@article{BrennerBDDC_2013,
author = {Brenner, Susanne C. and Park, Eun-Hee and Sung, Li-Yeng},
title = {A balancing domain decomposition by constraints preconditioner for a weakly over-penalized symmetric interior penalty method},
journal = {Numerical Linear Algebra with Applications},
volume = {20},
number = {3},
pages = {472-491},
doi = {https://doi.org/10.1002/nla.1838},
url = {https://onlinelibrary.wiley.com/doi/abs/10.1002/nla.1838},
eprint = {https://onlinelibrary.wiley.com/doi/pdf/10.1002/nla.1838},
year = {2013}
}
\bibliographystyle{tmlr}
%
\end{document}